\documentclass[sigconf]{acmart}
\usepackage{algorithmic}
\usepackage{algorithm}
\usepackage{multirow}
\usepackage{mhchem}
\usepackage{xcolor}
\usepackage{color}
\usepackage{colortbl}
\definecolor{c1}{HTML}{95bddc}
\definecolor{c2}{HTML}{FF4500}
\definecolor{c3}{HTML}{92CDDC}
\definecolor{c4}{HTML}{336626}
\definecolor{c5}{HTML}{FFB400}

\AtBeginDocument{%
  }

\copyrightyear{2026}
\acmYear{2026}
\setcopyright{cc}
\setcctype{by}
\acmConference[KDD '26]{Proceedings of the 32nd ACM SIGKDD Conference on Knowledge Discovery and Data Mining V.2}{August 09--13, 2026}{Jeju Island, Republic of Korea}
\acmBooktitle{Proceedings of the 32nd ACM SIGKDD Conference on Knowledge Discovery and Data Mining V.2 (KDD '26), August 09--13, 2026, Jeju Island, Republic of Korea}
\acmDOI{10.1145/3770855.3818850}
\acmISBN{979-8-4007-2259-2/2026/08}

\begin{document}

\title{Representational Alignment with Chemical Induced Fit for Molecular Relational Learning}

\author{Peiliang Zhang}
\email{cheungbl@ieee.org}
\affiliation{%
\institution{Wuhan University of Technology}
  \city{Wuhan}
  % \state{Hubei}
  \country{China}
}
\affiliation{%
  \institution{Yonsei University}
  \city{Seoul}
  \country{Republic of Korea}
}

\author{Jingling Yuan}
\email{yjl@whut.edu.cn}
\authornote{Corresponding author.}
\affiliation{%
\institution{Hubei Key Laboratory of Transportation Internet of Things}
  \city{Wuhan}
  \state{Hubei}
  \country{China}
}
\affiliation{%
\institution{Wuhan University of Technology}
  \city{Wuhan}
  \state{Hubei}
  \country{China}
}

\author{Qing Xie}
\email{felixxq@whut.edu.cn}
\affiliation{%
\institution{Wuhan University of Technology}
  \city{Wuhan}
  \state{Hubei}
  \country{China}
}

% \author{Qing Xie}
% \email{felixxq@whut.edu.cn}
% \affiliation{%
%   \institution{Wuhan University of Technology}
% }

\author{Yongjun Zhu}
\email{zhu@yonsei.ac.kr}
\affiliation{%
  \institution{Yonsei University}
  \city{Seoul}
  \country{Republic of Korea}
}

\author{Chao Che}
\email{chechao@gmail.com}
\affiliation{%
  \institution{Dalian University}
  \city{Dalian}
  \state{Liaoning}
  \country{China}
}

\author{Lin Li}
\email{cathylilin@whut.edu.cn}
\affiliation{%
\institution{Wuhan University of Technology}
  \city{Wuhan}
  \state{Hubei}
  \country{China}
}

\renewcommand{\shortauthors}{Peiliang Zhang et al.}

\begin{abstract}
Molecular Relational Learning (MRL) is widely applied in the natural sciences to predict relationships between molecular pairs by extracting structural features. The representational similarity between substructure pairs determines the functional compatibility of molecular binding sites. Nevertheless, aligning substructure representations with attention mechanisms lacks guidance from chemical knowledge, resulting in unstable model performance in chemical space (\textit{e.g.}, functional group, scaffold) shifted data. With theoretical justification, we propose the \textbf{Re}presentational \textbf{Align}ment with Chemical Induced \textbf{Fit} (ReAlignFit) to enhance the stability of MRL. ReAlignFit dynamically aligns substructure representation in MRL by introducing chemical Induced Fit-based inductive bias. In the induction process, we design a Bias Correction Function based on substructure edge reconstruction to align representations between substructure pairs by simulating chemical conformational changes (dynamic combination of substructures). ReAlignFit further integrates the Subgraph Information Bottleneck during the fitting process to refine and optimize substructure pairs exhibiting high chemical functional compatibility, leveraging them to generate molecular embeddings. Experimental results on nine datasets demonstrate that ReAlignFit outperforms state-of-the-art models in two tasks and significantly enhances model’s stability in both rule-shifted and scaffold-shifted data distributions.
\end{abstract}


\begin{CCSXML}
<ccs2012>
   <concept>
       <concept_id>10010405.10010444.10010450</concept_id>
       <concept_desc>Applied computing~Bioinformatics</concept_desc>
       <concept_significance>500</concept_significance>
       </concept>
 </ccs2012>
\end{CCSXML}

\ccsdesc[500]{Applied computing~Bioinformatics}



\keywords{Molecular Relational Learning, Representational Alignment, Chemical Induced Fit, Model Stability}


\maketitle

\section{Introduction}
Molecular Relational Learning (MRL) predicts the interactions between molecular pairs by mining features and properties~\cite{TNNLS2,zhang2026prototype,lee2023shiftKDD}. MRL has garnered significant attention in the natural science research due to its applications in new material design and drug discovery~\cite{tkdeGIBB,chi,TNNLS3}. 
Molecular structure representation-based methods have advantages in MRL and demonstrate satisfactory performance in downstream tasks~\cite{wang2025image,wang2024kdd,TNNLS1}. Therefore, accurately representing molecular features becomes critical in MRL.

The functional compatibility of binding sites is an essential determinant of molecular relationships~\cite{koshland1995key,mcgibbon2024intuition,xia2023understanding}.
Recent research primarily quantifies functional compatibility by calculating similarity between substructure representations, focusing on molecular representational alignment by attention-based inductive bias~\cite{maboudi2024retuning,seo2024selfKDD,xia2023understanding}.
The molecular interaction is regarded as an essential source of inductive bias, which aligns the properties of paired molecules with the representations of the molecular core features by modeling the chemical reactions between the molecules. 
The core challenge lies in effectively computing inductive bias within molecular relationships. 
Existing methods have explored molecular-level~\cite{zhao2025evidential}, substructure-level~\cite{lee2023cgibICML}, and hybrid strategy alignment~\cite{dsn}.

Nevertheless, applying inductive bias-based representational alignment to MRL involves two key considerations: (1) \textbf{Guidance from Domain Knowledge}: Molecular feature representations are frequently influenced by domain knowledge in chemistry or biology~\cite{xia2023understanding,lu2025dtiam,nc}.
Changes in adjacent atoms or scaffolds within the chemical space influence the chemical properties of molecular substructures~\cite{zhang2026kdd,li2023reaction,yang2023molerecwww}. 
% Substructures within molecules are susceptible to influence by surrounding atoms and scaffolds, which may change some chemical properties~\cite{li2023reaction,yang2023molerecwww}. 
Most of the existing representational alignment methods~\cite{saddi,yang2024interaction,lee2023shiftKDD} predominantly compute inductive bias using attention mechanisms, leading the results to reflect statistical correlations and overlook other active atoms that may affect the properties of these substructures. (2) \textbf{Dynamic Adaptability of Inductive Bias}: The substructures (functional groups within molecules) involved in chemical reactions are closely related to their paired substructures \cite{mak2024artificial,lu2025dtiam,lu2025dtiam}, meaning that substructures involved in different chemical reactions within the same molecule may vary. 
Structural shifts in functional groups and scaffolds may induce dynamic changes in the importance patterns of substructures in chemical reactions.
Static inductive bias based on attention mechanisms~\cite{miracle,zhang2024key,su2024dualAAAI} tends to concentrate weights on certain substructures with specific features, failing to capture dynamic changes such as the adaptive adjustment of substructures in MRL.

To address these challenges, this study aims to enhance model stability by integrating chemical Induced Fit theory into substructure representational dynamic alignment. We first theoretically demonstrate that aligning information between core substructure pairs facilitates stable MRL. With theoretical justification, we propose the \textbf{Re}presentational \textbf{Align}ment with Chemical Induced \textbf{Fit} (ReAlignFit) to improve MRL stability. ReAlignFit generates substructure embedding through the GNN encoder. Inspired by Induced Fit theory~\cite{koshland1995key}, we design the Dynamic Representational Alignment Module (DRAM) with substructure edge reconstruction, whose core is the Bias Correction Function (BCF). BCF simulates dynamic conformational changes (\textit{i.e.}, dynamic adjustments among substructures) during Induced Fit by a self-supervised approach. Combined with the Subgraph Information Bottleneck (S-GIB), DRAM aligns core substructure pairs for potential compatibility in chemical function. ReAlignFit integrates domain knowledge while mitigating errors in identifying core substructures and aligning information caused by spurious attention. Ultimately, we train ReAlignFit by synergistically optimizing molecular representational alignment's confusion loss and task prediction loss.
Experimental results demonstrate that ReAlignFit achieves state-of-the-art predictive performance.
\begin{itemize}
    \item We propose ReAlignFit, which introduces domain knowledge into representational alignment for stable MRL. To the best of our literature review, it is the first work to explore the dynamic representational alignment of substructures across different chemical reactions.
    \item With theoretical justification, we provide a formal certification for the loss function design which aligns representations between substructures by minimizing differences among core substructures.
    \item Experiments on two tasks across nine datasets demonstrate that ReAlignFit outperforms 13 state-of-the-art models and improves model stability in data shifted distributions.
\end{itemize}

\section{Preliminaries}
\label{Preliminaries}
In this section, we illustrate how the attention mechanism-based inductive bias affects the stability of MRL with specific examples. We formally describe stable MRL and present a theoretical analysis to identify feasible solutions to improve MRL stability.

\begin{figure}[t]
\centering
\includegraphics[width=0.9\columnwidth]{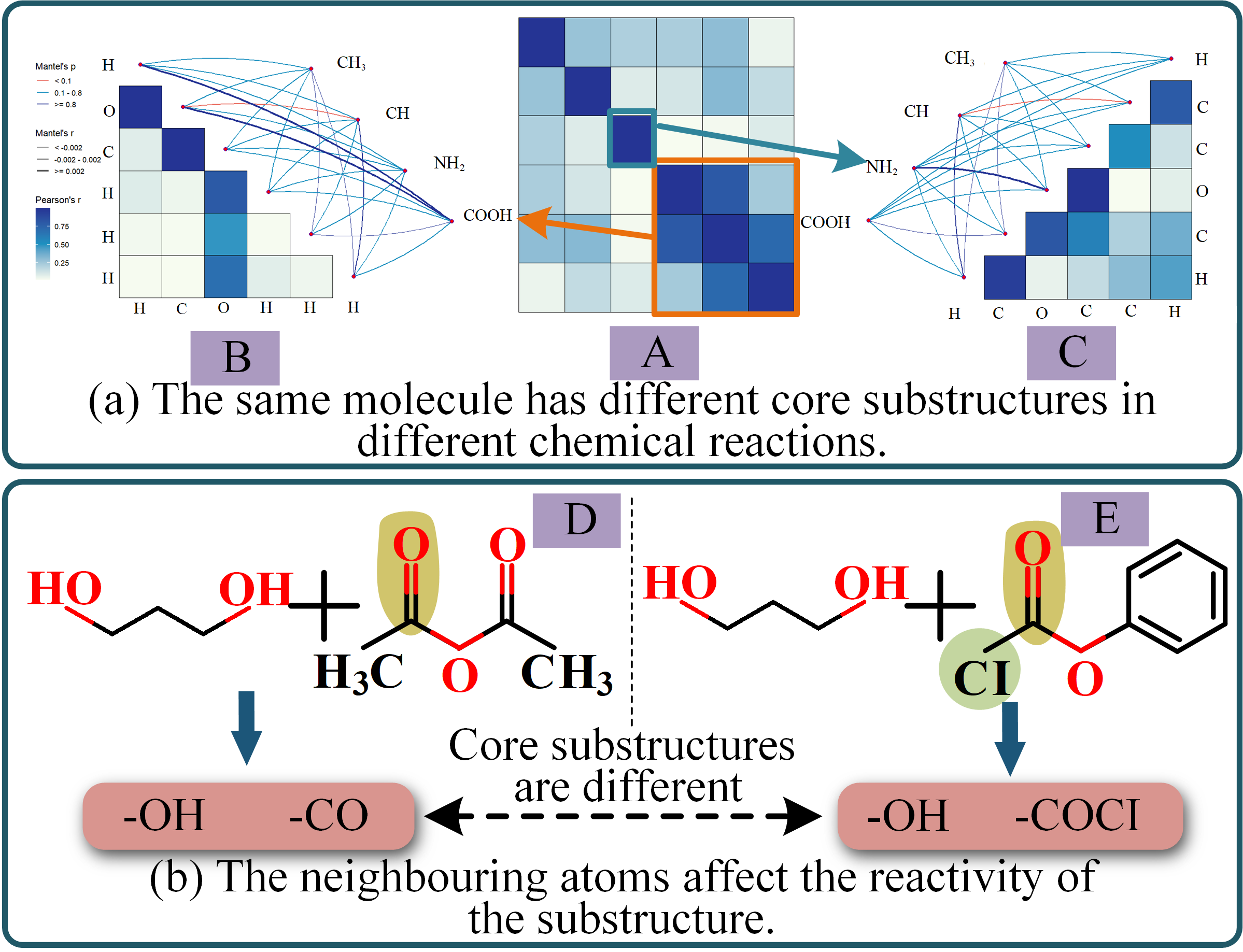}
\caption{The motivating example. (a) When molecule A reacts with molecules B and C, the core substructures are \textcolor{c2}{\ce{-COOH}} and \textcolor{blue}{\ce{-NH2}}, respectively. (b) The properties of \textcolor{c5}{\ce{-CO}} within molecular E are influenced by surrounding reactive atoms \textcolor{c4}{\ce{-CI}}, leading to changes in its behavior.}
\label{fig1}
\end{figure}

\subsection{Motivating Example}
\label{Motivating Example}
The inherent biases and errors in the inductive bias introduced by attention mechanisms~\cite{cao2024towards} are critical factors affecting the stability of MRL. In this subsection, we explain in further detail how such errors in inductive bias affect the representational alignment between molecular pairs and the stability of MRL with the example in Fig. \ref{fig1}.

We will explain the impact of inductive bias errors on the representational alignment between substructure pairs by considering the dynamic adaptation of the induction bias and the guidance of chemical domain knowledge. 
(1) \textbf{The impact of dynamic adaptability of inductive bias on representational alignment}: The core substructure determines the reaction may change depending on the paired molecule. In Fig. \ref{fig1}\textcolor{blue}{(a)}, when molecule A reacts with B, the reactive substructure in A is \textcolor{c2}{\ce{-COOH}}, while with C, it is \textcolor{blue}{\ce{-NH2}}. The scaffold of molecule B is \ce{-OH}, while that of molecule C is \ce{C=O}. This variation in scaffolds results in different reactive substructures in molecule A. However, the \textcolor{c2}{\ce{-COOH}} is slightly more reactive than \textcolor{blue}{\ce{-NH2}} in chemical reactions. The data co-occurrence model of the attentional mechanism causes the inductive bias to overly focus on \textcolor{c2}{\ce{-COOH}} and ignore the \textcolor{blue}{\ce{-NH2}}, which are the true drivers of chemical reactions. This phenomenon indicates that attention-based inductive bias lacks dynamic adaptability to paired molecules in MRL, making it difficult to identify and align the representations of core substructures involved in chemical reactions.
(2) \textbf{The guidance of domain knowledge on representational alignment}: The properties of molecular substructures are susceptible to influence by surrounding atoms. As shown in Fig. \ref{fig1}\textcolor{blue}{(b)}, the strong electron-adsorption group \textcolor{c4}{\ce{-CI}} in molecule E enhances the reactivity of \textcolor{c5}{\ce{-CO}}. In the esterification reaction, the \ce{-COCI} in molecule E will replace \textcolor{c5}{\ce{-CO}} and react with the hydroxyl group in alcohols. 
Substructure representation alignment without domain knowledge guidance (e.g., inductive bias-based attention or feature concatenation) usually assigns higher weights to more frequently occurring \textcolor{c5}{\ce{-CO}}, hindering the chemical consistency of core substructure representation alignment.

In summary, the core substructures that drive chemical reactions are critical factors influencing performance fluctuations of MRL models in distribution-shifted data. In contrast, existing methods lack the dynamic adaptability needed in capturing and representing these substructures.
Therefore, a key challenge in enhancing the stability lies in how to dynamically align the representations of core substructures under the guidance of chemical domain knowledge.

\subsection{Stable Molecular Relational Learning}
\label{Stable Molecular Relational Learning}
\subsubsection{Molecular Representation and MRL}
We first introduce the problem definitions of molecular representation and MRL in detail.
For any molecule $\mathcal{G}$, it can be represented as $\mathcal{G}=(V,\mathcal{E},\mathcal{X},A)$. 
Here, $V=\{{{v}_{1}},{{v}_{2}},\cdots ,{{v}_{N}}\}$ denotes the set of nodes. $\mathcal{E}\in N\times N$ represents the connections between atoms within the molecule, which is closely related to the adjacency matrix $A$. If $({{v}_{i}},{{v}_{j}})\in \mathcal{E}$, then ${{A}_{ij}}=1$; otherwise, ${{A}_{ij}}=0$. $\mathcal{X} \in {\mathbb{R}^{B\times N}}$ is the feature matrix, consisting of the atom feature representations. For a given molecular pair $({\mathcal{G}_{x}}, {\mathcal{Y}_{xy}}, {\mathcal{G}_{y}}) $, the objective of MRL is to construct a model $\mathcal{F}_{MRL}=(\mathcal{F}_{End},\mathcal{F}_{Pred})$ that generates molecular embedding representations $\mathcal{H}_x$ and $\mathcal{H}_y$ through the encoder $\mathcal{F}_{End}$, and predicts the relationship $\mathcal
{\hat{Y}}_{xy}$ between moleculars with the classifier $\mathcal{F}_{Pred}$.

\subsubsection{Stable MRL}
We further give a detailed definition of Stable MRL based on MRL. Given the dataset $\mathcal{D}=\{\mathcal{D}_{tra},\mathcal{D}_{val},\mathcal{D}_{tes}\}$, $\mathcal{D}_{tra}, \mathcal{D}_{val}, \mathcal{D}_{tes}$ are the training set, validation set, and test set, respectively. In the case of data distribution bias, the data distribution in $\mathcal{D}_{tra}, \mathcal{D}_{val}, \mathcal{D}_{tes}$ are different from each other. Formally, $\mathcal{S}(\mathcal{D}_{tra}) \neq \mathcal{S}(\mathcal{D}_{val})$, $\mathcal{S}(\mathcal{D}_{tra}) \neq \mathcal{S}(\mathcal{D}_{tes})$, $\mathcal{S}(\mathcal{D}_{val}) \neq \mathcal{S}(\mathcal{D}_{tes})$. The specific setup and partitioning of $\mathcal{D}_{tra},\mathcal{D}_{val},\mathcal{D}_{tes}$ will be described in detail in Section \ref{Experimental Setup}. Stable MRL aims to train a model $\mathcal{F}_{MRL}^{S}=(\mathcal{F}_{End}^{tra},\mathcal{F}_{Pred}^{val},\mathcal{F}_{Pred}^{tes})$ using $\mathcal{D}_{tra}$ and $\mathcal{D}_{val}$ that generalizes well to $\mathcal{D}_{tes}$. The model $\mathcal{F}_{MRL}$ should maintain relatively stable predictive performance in different distribution shifts.
ReAlignFit aims to learn stable molecular representations consisting of core substructures, thereby improving the model's performance.

\subsection{Theoretical Analysis of Stable MRL}
\label{Theoretical Analysis of Stable MRL}
Since modeling molecular relationships solely from the data perspective often leads to instability in the model, we introduce the Induced Fit theory in chemistry to identify key factors affecting the stability of MRL during theoretical analysis.

The Induced Fit theory describes the dynamic mechanism of specific molecular binding. It emphasizes that the matching of binding sites (\textit{i.e.}, representation similarity between paired substructures) is critical for enhancing binding stability \cite{koshland1995key,stiller2022structure}. Inspired by the Induced Fit theory, we attempt to analyze the contribution of substructure-pair matching to MRL performance stabilization on the theoretical level.
\begin{theorem}
\label{theorem 1}
    \textit{Given the molecular pair $({\mathcal{G}_{x}},{\mathcal{G}_{y}})$ and the prediction target $\mathcal{Y}$, where the substructure ${\mathcal{G}^{s}}$ of $\mathcal{G}$ consists of core substructure ${\mathcal{G}^{c}}$ and confounding substructure ${\mathcal{G}^{n}}$. For $\forall$ ${\mathcal{G}_{x}},{\mathcal{G}_{y}}\in \mathcal{G}$, according to the law of conditional probability, $\mathcal{P}({\mathcal{G}_{x}},{\mathcal{G}_{y}};\mathcal{Y})\geq \mathcal{P}({\mathcal{G}^{c}};\mathcal{Y}|{\mathcal{G}^{n}})$. Furthermore, considering the correlation between ${\mathcal{G}^{c}}$ and ${\mathcal{G}^{n}}$, if there exists a minimal value $\varepsilon $ such that:}
\begin{equation}
\label{eq:theorem}
\left| \mathcal{P}({\mathcal{G}_{x}},{\mathcal{G}_{y}};\mathcal{Y})\!-\!\mathcal{P}(\mathcal{G}_x^c,\mathcal{G}^c_y;\mathcal{Y})\!+\!\mathcal{P}(\mathcal{G}_x^c;\mathcal{G}_x^n)\!+\!\mathcal{P}(\mathcal{G}_y^c;\mathcal{G}_y^n) \right| \! \le \! \varepsilon 
\end{equation}
\textit{where $\mathcal{P}({\mathcal{G}_{x}},{\mathcal{G}_{y}};\mathcal{Y})$ is the true probability between molecular pair and the prediction target. 
$\mathcal{P}(\mathcal{G}_{x}^{c},\mathcal{G}_{y}^{c};\mathcal{Y})$ is interaction probability between core substructure captured by model learning and prediction target called learning probability. $\mathcal{P}({\mathcal{G}^{c}};{\mathcal{G}^{n}})$ is the confounding probabilities between core and confounding substructures.}
\end{theorem}

The $\varepsilon $ measures the similarity between true probability, learning probability, and confounding probability. The discrepancy between true and learned probabilities is derived from the task's prediction loss, while the degree of calibration between substructure representations quantifies the confusion probability. 
Theorem \ref{theorem 1} demonstrates that when $\eta $ is sufficiently small, meaning the prediction loss and confusion probability are small enough, the relationships between molecular pairs can be stably represented. Motivated by Theorem \ref{theorem 1}, we optimize model learning by minimizing prediction loss and confusion probability. This approach guides us in identifying core substructures by substructure representational alignment while reducing the impact of non-core substructures on molecular representations, thereby improving the stability of MRL. The proof of Theorem \ref{theorem 1} is provided in Appendix \ref{The Proof of Theorem 1}.

\begin{figure*}[htpb]
\centering
\includegraphics[width=\textwidth]{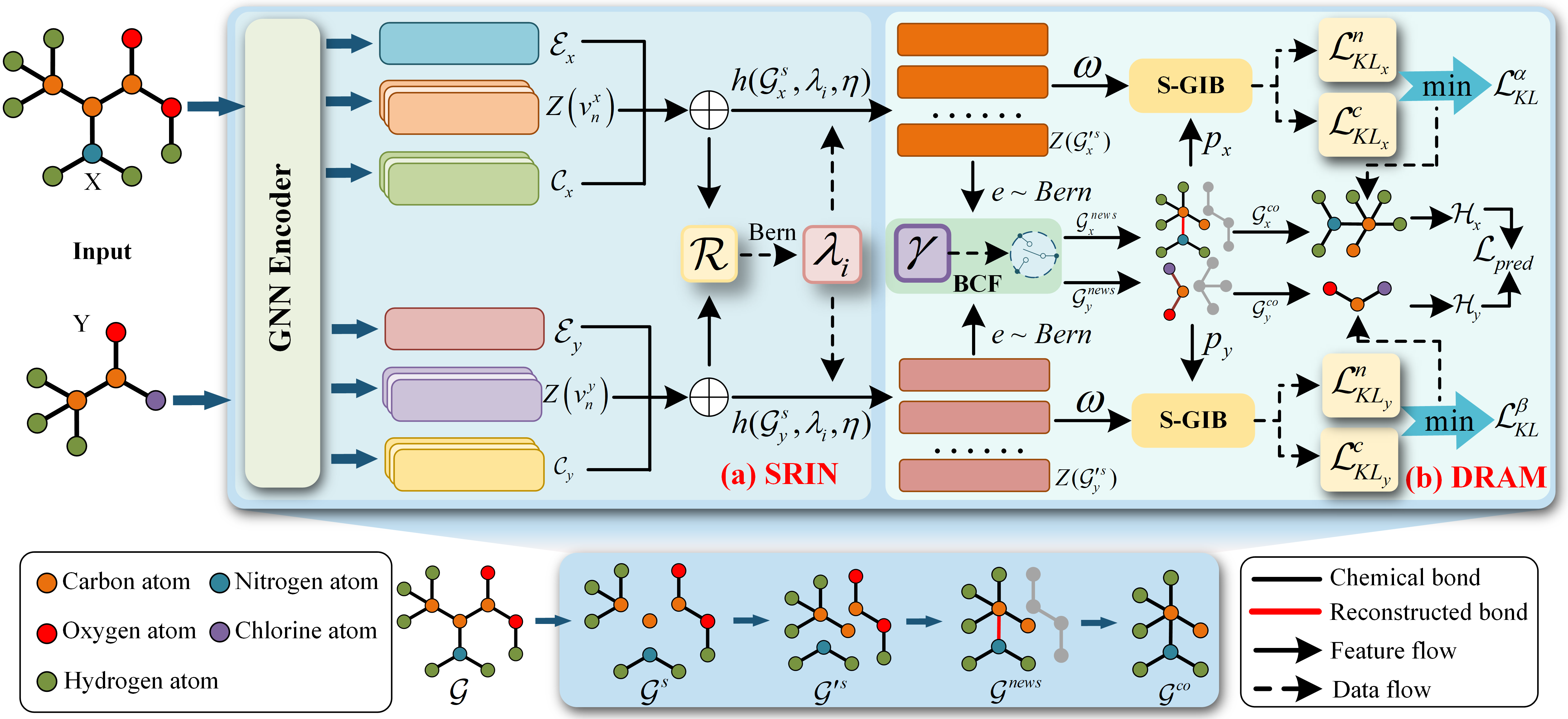} 
\caption{The model structure of ReAlignFit. (a)
SRIN generates substructure representations. (b) DRAM aligns and optimizes the core substructure representations to generate stable representations of molecules.}
\label{model}
\end{figure*}

\section{Methodology}
\label{Methodology}
Inspired by Theorem \ref{theorem 1}, we propose ReAlignFit to enhance the stability of MRL by simulating the dynamic mechanism of molecular binding, as shown in Fig. \ref{model}. Based on Substructure Representation-based Interaction Network (SRIN) in Fig. \ref{model}\textcolor{blue}{(a)}, we design the Dynamic Representational Alignment Module (DRAM), as shown in Fig. \ref{model}\textcolor{blue}{(b)}, which employs BCF and S-GIB to align the core substructure representations (Section \ref{Representational Alignment with Induced Fit}). 
Additionally, we present the transformation and solution process of Subgraph Information Bottleneck (S-GIB) (Section \ref{Synergistic Optimization Loss Function for ReAlignFit}) and analyze the computational complexity of ReAlignFit (Section \ref{Model Analysis}).

\subsection{Representational Alignment with Induced Fit}
\label{Representational Alignment with Induced Fit}
The Induced Fit theory suggests that chemical reactions between molecules are realized by combining the pre-adaptation of substructures and dynamic adjustments during binding~\cite{koshland1995key}. Guided by this mechanism, we design SRIN to simulate the pre-adaptation process. In DRAM, BCF simulates the molecular induction process to align substructure representations, while S-GIB identifies and refines substructures with high chemical functional compatibility.

\subsubsection{Substructure Representation-based Interaction Network}
\label{Substructure Representation-based Interaction Network}
Considering the irregularity of molecular spatial topologies and the effectiveness of GNN in dealing with topological features~\cite{du2024mmgnnIJCAI,zhang2024heterogeneousIJCAI}, ReAlignFit employs a GNN encoder with an adjacency aggregation strategy to generate embeddings for irregular substructures. 

Initially, molecule $\mathcal{G}$ represented by SMILES~\cite{smiles} are converted into GNN-compatible representations by RDKit~\cite{brown2015silico}. Subsequently, we use neighborhood feature weighting consisting of adjacency matrix $\mathcal{E}$, node feature representation $\mathcal{Z}(v_n)$, and neighborhood structural coefficient $\mathcal{C}$ as the input for irregular substructure generation. The $\mathcal{Z}^{l+1}(v_n)$ in $(l+1)$th layer and $\mathcal{C}$ are defined as follows:
\begin{equation}
\label{node}
\begin{aligned}
    \mathcal{Z}^{l+1}&(v_n)=\sum \nolimits_{v_u \in \mathcal{N}(v_n)}( W_u \mathcal{Z}^{l}(v_u) + W_{n} \mathcal{Z}^{l}(v_n))\\
    &\mathcal{C}=\frac{\sigma (\mathcal{Z}(v_n))\cdot \log \sigma (\mathcal{Z}(v_n))}{\sum \nolimits_{v_u \in \mathcal{N}(v_n)}\sigma (\mathcal{Z}(v_u))\cdot \log \sigma (\mathcal{Z}(v_u))}
\end{aligned}
\end{equation}
where $\mathcal{N}(v_n)$ denotes the set of neighbor nodes of $v_n$. $W$ is the weight matrix and $\sigma$ is the activation function. $\mathcal{Z}(*)$ is the embedding representation obtained from the GNN encoder.
For molecule $\mathcal{G}$, its irregular substructure $\mathcal{Z}(\mathcal{G}^s)$ is derived by aggregating the central node $v_n$ and its $K$-hop neighbors $v_n^k$, weighted by $\mathcal{C}$:
\begin{equation}
    \label{irr}
    \mathcal{Z}({\mathcal{G}^s}) = \sum \nolimits_{k=1}^K \sum \nolimits_{{{v}_{u}}\in v_{n}^{k}}(\mathcal{Z}(v_n)||{{\mathcal{C}}} \cdot \mathcal{Z}(v_u))
\end{equation}

In the subsequent step, we calculate the interaction probability between substructures using Eq (\ref{eq:interpro}) to simulate the pre-adaptation process of substructures. We further optimize the molecular substructure representations by eliminating noise with Eq (\ref{eq:gen}) results. The substructure interaction probability for pre-adaptation is:
\begin{equation}
\label{eq:interpro}
\mathcal{R}_i\!=\!\!\!\!\!\!\sum \limits_{\mathcal{G}_{y}^{s_j}\in \mathcal{G}_{y}^{s}} \!\!\!{\sigma(\mathcal{Z}(\mathcal{G}_{x}^{s_i}),\mathcal{Z}(\mathcal{G}_{y}^{s_j}))\!+\!\frac{1}{J\!\!-\!\!1}\!\!\!\!\! \sum \limits_{\mathcal{G}_{y}^{s_k} \neq \mathcal{G}_{y}^{s_j}}\!\!\!\! {\sigma(\mathcal{Z}(\mathcal{G}_{x}^{s_i}),\mathcal{Z}(\mathcal{G}_{y}^{s_k}))}}
\end{equation}
where $\sigma ()$ is a probability function. $\mathcal{G}_{x}^{s_i}$ and $\mathcal{G}_{y}^{s_j}$ denote the substructures of the paired molecules $(\mathcal{G}_{x},\mathcal{G}_{y})$. $\mathcal{G}_{y}^{s}$ represents the substructure set of molecule $\mathcal{G}_{y}$, containing $J$ elements. ReAlignFit eliminates unimportant substructures (i.e., those with lower $\mathcal{R}_i$) via the $\mathcal{R}_i$ and the noise rejection function $\lambda$ to control the effect of confounding information on substructure representations thereby generating the optimized substructure representation $\mathcal{Z}({\mathcal{G}}'^s)$:
\begin{equation}
\label{eq:gen}
  \mathcal{Z}({\mathcal{G}}'^s)=h({\mathcal{G}}^s,\lambda_i,\eta)={{\lambda }_{i}}\mathcal{Z}({\mathcal{G}^{s_i}})+(1-{{\lambda }_{i}})\eta
\end{equation}
where ${{\mu }_{\mathcal{Z}(\mathcal{G}_{{}}^{s})}}$ and $\sigma _{\mathcal{Z}(\mathcal{G}_{{}}^{s})}^{2}$ are mean and variance of $\mathcal{Z}(\mathcal{G}_{{}}^{s})$, respectively. The refined $\mathcal{Z}({\mathcal{G}}'^s_x)$ and $\mathcal{Z}({\mathcal{G}}'^s_y)$ are utilized to dynamic representational alignment. ${\lambda }_{i}$ is obtained from a Bernoulli sampling consisting of $\mathcal{R}_i$ and $\mathcal{Z}(\mathcal{G}^{s})$.

\subsubsection{Dynamic Representational Alignment Module}
\label{Dynamic Representation Alignment with Bias Correction Function} 
Representational alignment methods that introduce inductive bias via attention mechanisms rely on static combinations between substructures, overlooking the dynamic adjustments in reactions. This limits their ability to capture core substructures driving different chemical reactions. 
The Induced Fit theory highlights conformational changes of paired substructures, \textit{i.e.}, the dynamic variations of substructure topologies and binding sites during molecular interactions. 
Therefore, we design the BCF to dynamically adjust substructure binding sites and representational alignment while integrating the S-GIB to identify highly functional compatible substructures.

Considering that GNN-based molecular representation methods may overlook the connecting mechanisms of the chemical bonds, we employ a substructure edge reconstruction strategy in DRAM to represent substructure information more accurately. 
Specifically, we remove the edges between all substructures $\mathcal{G}^s$ in $\mathcal{G}$ at the initial stage. DRAM computes the Bernoulli distribution $e_{ik}$ between substructures $\mathcal{G}^{s_i}$ and $\mathcal{G}^{s_k}$ by edge sampling, and reconstruct the edges with $e_{ik}=1$. $e_{ik} \sim \text{Bernoulli}(\theta_{ik})$, where $\theta_{i,k}$ is a Gaussian kernel function parameterized by $ \tau\!\!=\!\! ||\!\mathcal{Z}(\mathcal{G}^{s_i}\!)\!-\!\!\mathcal{Z}(\!\mathcal{G}^{s_k}\!)||^2_2 / \!\!\sqrt{2}$.
% where $\theta_{ik}$ is computed using a Gaussian Kernel function. 
The new substructure obtained in reconstruction is denoted as $\mathcal{G}^{news}$.
\begin{equation}
\theta_{ik}=\frac{\exp (-||\mathcal{Z}(\mathcal{G}^{s_i})-\mathcal{Z}(\mathcal{G}^{s_k})||^2/ 2\tau^2)}{\sum_{\mathcal{G}^s \in \mathcal{G}}\exp (-||\mathcal{Z}(\mathcal{G}^{s})-\mathcal{Z}(\mathcal{G}^{s_k})||^2 / 2\tau^2)}
\end{equation}

After edge reconstruction, we design the Bias Correction Function $\gamma$ to quantify the degree of alignment between substructures before and after reconstruction. 
\begin{equation}
\label{align}
    \gamma \!=\!\sum \limits_{j\leq J}\frac{2\left\| \sigma(\mathcal{Z}(\mathcal{G}_{x}^{news}),\mathcal{Z}(\mathcal{G}_{y}^{s_j})) \right\|_{2}^{2}}{J(\left\| \sigma(\!\mathcal{Z}(\mathcal{G}_{x}^{s_i}),\!\mathcal{Z}(\mathcal{G}_{y}^{s_j})) \right\|_{2}^{2}\!+\!\left\| \sigma(\!\mathcal{Z}(\mathcal{G}_{x}^{s_k}),\mathcal{Z}(\mathcal{G}_{y}^{s_j})) \right\|_{2}^{2})}
\end{equation}
where $\mathcal{G}_{x}^{news}$ is reconstructed from the substructures $\mathcal{G}_{x}^{s_i}$ and $\mathcal{G}_{x}^{s_k}$ of $\mathcal{G}_x$. $\mathcal{G}_{y}^{s_j}$ represents the substructure of $\mathcal{G}_y$ paired with $\mathcal{G}_x$. 

In the Induced Fit theory, conformational changes aim to achieve functional compatibility at binding sites, which can be approximately quantified through similarity in vector space~\cite{stiller2022structure}. Therefore, we employ cosine similarity to compute the components of $\gamma$, quantifying the chemical functional compatibility.

DRAM dynamically determines whether to perform edge reconstruction and align substructure representation.  based on $\gamma$. The substructure representation after dynamic adjustment is
\begin{equation}
    \mathcal{G}^{c}_x =
\begin{cases} 
  \text{if } \gamma \geq 1, & \mathcal{G}_{x}^{news} \\
  \text{if } \gamma < 1, & 
  \begin{cases}
      \text{if } \sigma_{ij} \geq \sigma_{kj}, & \mathcal{G}_{x}^{s_i}\\
      \text{if } \sigma_{ij} < \sigma_{kj}, & \mathcal{G}_{x}^{s_k}\\
  \end{cases} \\
\end{cases}
\label{if}
\end{equation}
where $\sigma_{*j}=\left\| \sigma(\mathcal{Z}(\mathcal{G}_{x}^{s_*}),\mathcal{Z}(\mathcal{G}_{y}^{s_j})) \right\|_{2}^{2}$. The substructure $\mathcal{G}^{c}_y$ after dynamic adjustment of $\mathcal{G}_y$ can be computed similarly.

Considering the advantages of the Graph Information Bottleneck (GIB) in graph optimization and the specific requirements of MRL tasks~\cite{hu2024GIBsurvey,zhang2025iterative}, we further extend the optimization objective of GIB to the subgraph level for core substructure selection and optimization. The refined objective is defined as follows:
\begin{definition}
    \label{Definition 1}
    (\textit{\textbf{S-GIB}}) \textit{Given a set of graphs and their interaction relationships $({\mathcal{G}_{x}}, \mathcal{Y}, {\mathcal{G}_{y}})$, $(\mathcal{G}^{c}_x, \mathcal{G}^{c}_y)$ and $(\mathcal{G}^{n}_x, \mathcal{G}^{n}_y)$ are the core subgraph pairs and confounding subgraph pairs of $({\mathcal{G}_{x}}, {\mathcal{G}_{y}})$ and $\mathcal{Y}$, respectively. The subgraph ${\mathcal{G}^{s}} = \{\mathcal{G}^{c}, \mathcal{G}^{n}\}$. According to the minimal sufficient principle of mutual information, the objective is}
\begin{equation}
\label{eq:CSGIBnew}
\mathcal{G}^{co}\!=\!{\mathop{\arg \min }}(\alpha \mathcal{I}(\mathcal{G}_{x}^{c},\mathcal{G}_{x}^{n})+\beta \mathcal{I}(\mathcal{G}_{y}^{c},\mathcal{G}_{y}^{n})-\mathcal{I}(\mathcal{Y};\mathcal{G}_{x}^{c},\!\mathcal{G}_{y}^{c}))
\end{equation}
\end{definition}
$-\mathcal{I}(\mathcal{Y};\mathcal{G}_{x}^{c},\mathcal{G}_{y}^{c})$ enables the model to fully learn the information of core subgraphs relevant to the prediction target. $\alpha \mathcal{I}(\mathcal{G}_{x}^{c},\mathcal{G}_{x}^{n})+\beta \mathcal{I}(\mathcal{G}_{y}^{c},\mathcal{G}_{y}^{n})$ minimizes the influence of confounding subgraphs on core subgraphs by eliminating confounding information.
The detailed solution process for S-GIB is presented in Section \ref{Synergistic Optimization Loss Function for ReAlignFit}.

We regard core substructures as core subgraphs in S-GIB and non-core substructures as confounding subgraphs. During the optimization of S-GIB, core substructures representing stable molecular representations are identified, while the influence of other substructures on molecular representations is minimized.
Through the synergy of Eqs (\ref{align})-(\ref{eq:CSGIBnew}), ReAlignFit dynamically selects and aligns the embedded representations of core substructures determining chemical reactions with full consideration of paired molecules.

The core substructure optimized by S-GIB is denoted as $\mathcal{G}^{co}$.
The final embedding representation $\mathcal{H}$ is generated by aggregating the core substructures of the molecules.
\begin{equation}
    \label{embedding}
    \mathcal{H}=\text{Readout}(\mathcal{Z}(\mathcal{G}^{co_1}) || \cdots || \mathcal{Z}(\mathcal{G}^{co_M})), M \ll  N
\end{equation}
where $M$ is the number of core substructures.

\subsection{Model Optimization and S-GIB Solution}
\label{Synergistic Optimization Loss Function for ReAlignFit}
We design the loss function $\mathcal{L}$ based on Theorem \ref{theorem 1} and demonstrate the transformation and solution process of S-GIB.

\subsubsection{Model Loss Function}
We design the model optimization objective $\mathcal{L}$ consistent with the derived conclusion in Theorem \ref{theorem 1}, which consists of prediction loss ${\mathcal{L}}_{pred}$ and calibration loss $\mathcal{L}_{KL}$.
\begin{equation}
\label{lossnum}
   \mathcal{L}= {\mathcal{L}}_{pred}+\alpha \mathcal{L}_{KL}^{\alpha }+\beta \mathcal{L}_{KL}^{\beta }
\end{equation}
where $\alpha$ and $\beta$ are hyperparameters that consistent with the settings in Eq~(\ref{eq:CSGIBnew}). To reduce the complexity of mutual information calculation, we transform Eq (\ref{eq:CSGIBnew}) as follows. 

\subsubsection{The Upper Bound of $\alpha \mathcal{I}(\mathcal{G}_{x}^{c},\mathcal{G}_{x}^{n}) + \beta \mathcal{I}(\mathcal{G}_{y}^{c},\mathcal{G}_{y}^{n})$ in Eq (\ref{eq:CSGIBnew})}
We utilize information-theoretic principles to derive the upper bounds of $\mathcal{I}(\mathcal{G}_{x}^{c},\mathcal{G}_{x}^{n})$ and $\mathcal{I}(\mathcal{G}_{y}^{c},\mathcal{G}_{y}^{n})$, respectively.

\begin{lemma}
    \label{Proposition 1}
    \textbf{(Upper bound of $\mathcal{I}(\mathcal{G}_{x}^{c},\mathcal{G}_{x}^{n})$)}
\textit{Since $\mathcal{G}^{c}_x$, $\mathcal{G}^{n}_x$ are subgraphs of ${G}_x$, according to the information transferability of Markov chain, we have
\begin{equation}
\label{eq:op1}
\begin{aligned}
\mathcal{I}(\mathcal{G}_{x}^{c},\!\mathcal{G}_{x}^{n})\!&\le\! \min (\mathcal{I}({\mathcal{G}^{c}_x},\mathcal{G}_x),\mathcal{I}({\mathcal{G}^{n}_x},\mathcal{G}_x))\\
&=\min \! (\!\!\int \!\!\!\!\! \int\!\!p(\!\mathcal{G}^c_x|{G}_x,\!\gamma)\! \log(\!\frac{p(\!\mathcal{G}^c_x|\mathcal{G}_x,\!\gamma)}{p(\mathcal{G}^c_x)}\!) d\mathcal{G}^c_xd\mathcal{G}_x, \\
&\quad  \!\!\!\int \!\!\!\!\! \int\!\!p(\mathcal{G}^n_x|\mathcal{G}_x,\gamma) \log(\frac{p(\mathcal{G}^n_x|\mathcal{G}_x,\gamma)}{p(\mathcal{G}^n_x)}) d\mathcal{G}^n_xd\mathcal{G}_x )\\
&:=\min(KL(p(\mathcal{G}^c_x|\mathcal{G}_x,\gamma)||p(\mathcal{G}^c_x)),\\
&\quad \quad \quad KL(p(\mathcal{G}^n_x|\mathcal{G}_x,\gamma)||p(\mathcal{G}^n_x))))\\
&=\min(\mathcal{L}_{KL_{x}}^{c},\mathcal{L}_{KL_{x}}^{n})
\end{aligned}
\end{equation}
where $p()$ is a posterior distribution function.}
\end{lemma}

Therefore, minimizing $\mathcal{L}_{KL}^{\alpha }=\min (\mathcal{L}_{K{{L}_{x}}}^{c},\mathcal{L}_{K{{L}_{x}}}^{n})$ provides an upper bound for the minimization of $\mathcal{I}(\mathcal{G}_{x}^{c},\mathcal{G}_{x}^{n})$. Similarly, the upper bound for minimizing $I(\mathcal{G}_{y}^{c},\mathcal{G}_{y}^{n})$ can be obtained by minimizing $\mathcal{L}_{KL}^{\beta }=\min (\mathcal{L}_{K{{L}_{y}}}^{c},\mathcal{L}_{K{{L}_{y}}}^{n})$.
The proof of Eq (\ref{eq:op1}) is in Appendix \ref{The proof of Eq op1}.

\textbf{The solution to Eq (\ref{eq:op1})}. To reduce the difficulty of computing mutual information in Eq (\ref{eq:op1}), we decompose probability distributions $p(\mathcal{G}^c|\mathcal{G},\gamma)$ and $p(\mathcal{G}^c)$ into variational approximation and multivariate Bernoulli distribution, respectively.
We redefine $p(\mathcal{G}^c)$ using the variational approximation $\omega(\mathcal{G}^c)$, represented by $e_{ik}$. 
$\omega(\mathcal{G}^c) =e_{ik}^{|\mathcal{G}^{s_i}|}(1-e_{ik}^{|\mathcal{G}^{s_i}|/|\mathcal{G}^{s_k}|})$. 
Inspired by \cite{TGNN}, we parameterize $p(\mathcal{G}^c_x|\mathcal{G}_x,\gamma)$ as a multivariate Bernoulli distribution:
\begin{equation}
    \label{multbern}
    p(\mathcal{G}^c_x|\mathcal{G}_x,\gamma) = \prod_{\mathcal{G}_x^{s_{i}} \in \mathcal{G}^c_x} p_x^c \cdot \prod_{\mathcal{G}_x^{s_{i}} \notin \mathcal{G}^c_x} (1 - p_x^c)
\end{equation}
where $p_x^c$ is the probability distribution given $\mathcal{G}_x$ and $\gamma$, which can be computed via $p(\mathcal{G}^c_x|\mathcal{G}^n_x,\mathcal{G}_x)$.
\begin{equation}
\label{pxc}
     p_x^c \!= \! p(\mathcal{G}^c_x|\mathcal{G}^n_x,\!\mathcal{G}_x) \!\! = \! \sigma \! \left( \mathcal{Z}(\mathcal{G}^c_x),\!(\!\mathcal{Z}(\mathcal{G}^{n_1}_x)||\cdots||\mathcal{Z}(\mathcal{G}^{n_i}_x\!)\!) \right)
\end{equation}
where $\sigma(*)$ is a sigmoid function. 
Therefore, the $\mathcal{L}_{KL_{x}}^{c} $ in Eq (\ref{eq:op1}) is calculated as follows:
\begin{equation}
    \label{klxc}
    \begin{aligned}
 \mathcal{L}_{KL_{x}}^{c}&=KL[p(\mathcal{G}^c_x|\mathcal{G}^n_x,\mathcal{G}_x)||\omega(\mathcal{G}^c)]\\
        &=\mathbb{E}_{\sim \mathcal{G}^c_x,\mathcal{G}_x} \!\!\! \left[\!\sum \limits p_x^c \log\frac{p_x^c}{e}\! +\! (1\!-\!p_x^c)\log\frac{1\!\!-\!\!p_x^c}{1\!\!-\!\!e}\!\right]
    \end{aligned}
\end{equation}

Similarly, we can obtain the $\mathcal{L}_{KL_{x}}^{n}$, $\mathcal{L}_{KL_{y}}^{c}$ and $\mathcal{L}_{KL_{y}}^{n}$.

\subsubsection{The Lower Bound of $\mathcal{I}(\mathcal{Y};\mathcal{G}_{x}^{c},\mathcal{G}_{y}^{c})$ in Eq (\ref{eq:CSGIBnew})}
We compute the lower bound of $\mathcal{I}(\mathcal{ Y};\mathcal{G}_{x}^{c},\mathcal{G}_{y}^{c})$ using $(\mathcal{G}_{x},\mathcal{G}_{y})$ and $\mathcal{Y}$.

\begin{lemma}
    \label{Proposition 2}
    \textbf{(Lower bound of $\mathcal{I}(\mathcal{Y};\mathcal{G}_{x}^{c},\mathcal{G}_{y}^{c})$)}
\textit{Given a graph pair $(\mathcal{G}_{x},\mathcal{G}_{y})$, its label information $\mathcal{Y}$, and the learned core subgraphs $(\mathcal{G}_{x}^{c},\mathcal{G}_{y}^{c})$, we have:
%给定一对图$（\mathcal{G}_{x}，\mathcal{G}_{y}）$，它的标签信息$\mathcal{Y}$，以及学习到核心子图的$(\mathcal{G}_{x}^{c},\mathcal{G}_{y}^{c})$，我们有
\begin{equation}
\label{eq:op2}
\begin{aligned}
\mathcal{I}_{ca}\!&=\mathcal{I}(\mathcal{Y};\mathcal{G}_{x}^{c},\mathcal{G}_{y}^{c})\\
&=H(\mathcal{Y})\!+\!\!\int\!\!{p(\mathcal{Y})\!\!\int\!\!\!\!\!\int\!\!{\mathcal{F}_{ca}(\mathcal{G}_{x}^{c},\mathcal{G}_{y}^{c}|\mathcal{Y})}d\mathcal{G}_{x}^{c}d\mathcal{G}_{y}^{c}d\mathcal{Y}}\\
&\ge \int{p(\mathcal{Y})(\int\!\!\!\!\!\int\!\!{\mathcal{F}_{ca}(\mathcal{G}_{x}^{c},\mathcal{G}_{y}^{c}|\mathcal{Y})}d{G}_{x}^{c}d\mathcal{G}_{y}^{c})d\mathcal{Y}}\\
& := \frac{1}{NM}\sum\nolimits_{n=1}^{N}{\sum\nolimits_{m=1}^{M}{\!q(\gamma (\mathcal{G}_{{{x}_{n}}}^{c},\mathcal{G}_{{{y}_{m}}}^{c}\!)|{{Y}}\!)}}\\
&=-{{\mathcal{L}}_{pred}}\\
\mathcal{F}&_{ca}(\mathcal{G}_{x}^{c},\mathcal{G}_{y}^{c}|\mathcal{Y})\!=\!q(\mathcal{G}_{x}^{c},\mathcal{G}_{y}^{c}|\mathcal{Y})\log(\frac{p(\mathcal{G}_{x}^{c},\mathcal{G}_{y}^{c}|\mathcal{Y})}{q(\mathcal{G}_{x}^{c},\mathcal{G}_{y}^{c})})
\end{aligned}
\end{equation}
where $q(\mathcal{G}_{x}^{c},\mathcal{G}_{y}^{c}|\mathcal{Y})$ is the variational approximation distribution used to approximate the posterior distribution $p()$. }
\end{lemma}

The Eq (\ref{eq:op2}) indicates that minimizing the prediction loss ${{\mathcal{L}}_{pred}}$ achieves the minimization of $-\mathcal{I}(\mathcal{Y};\mathcal{G}_{x}^{c},\mathcal{G}_{y}^{c})$. The proof of Eq (\ref{eq:op2}) is shown in Appendix \ref{The proof of Eq op2}.

\textbf{The solution to Eq (\ref{eq:op2})}.
The prediction term $q(\gamma (\mathcal{G}_{{{x}}}^{c},\mathcal{G}_{{{y}}}^{c})|{\mathcal{Y}})$ in Eq (\ref{eq:op2}) emphasizes leveraging the core substructure pair $(\mathcal{G}_{{x}}^{c},\mathcal{G}_{{y}}^{c})$ to predict the relationship between molecules $\mathcal{G}_{{x}}$ and $\mathcal{G}_{{y}}$.

According to Eq (\ref{embedding}), the embedding representations of molecules $\mathcal{G}_{{{x}}}$ and $\mathcal{G}_{{{y}}}$, composed of their core substructures, are denoted as $\mathcal{H}_x$ and $\mathcal{H}_y$, respectively. Finally, the prediction loss ${{\mathcal{L}}_{pred}}$ of ReAlignFit is defined as:
\begin{equation}
    \label{loss}
    {{\mathcal{L}}_{pred}}=\frac{-1}{MN}\mathbb{E}_{(\mathcal{G}_x,\mathcal{G}_y)\sim \mathcal{Y}}[\log (\sigma(\mathcal{H}_x,\mathcal{H}_y))|\mathcal{Y}]
\end{equation}

In MRL, ${{\mathcal{L}}_{pred}}$ can be chosen as the cross-entropy loss for Drug-Drug Interaction (DDI) prediction or the mean absolute error loss for Molecular Interaction (MI) prediction.
% , depending on the nature of the task.

\subsection{Model Analysis}
\label{Model Analysis}
We train ReAlignFit through iterative optimization between SRIN and DRAM. The pseudocode is provided in Algorithm \ref{alg:ReAlignFit}.
The computational complexity of ReAlignFit is mainly derived from iterative optimization. The complexity of SRIN is $\mathcal{O}(T_1\cdot L\cdot |N|^2)$, and that of DRAM is $\mathcal{O}(T_2 \cdot J^2)$. 
$T$ represents the number of iterations, $L$ is GNN layers. Consequently, the overall computational complexity of ReAlignFit is approximately $\mathcal{O}(T_1\cdot L\cdot |N|^2 +T_2 \cdot J^2)$. Since $J \ll N$, the added computational overhead is still manageable.

\begin{algorithm}[htpb]
\caption{ReAlignFit}\label{alg:ReAlignFit}
\textbf{Input}: Dataset $\mathcal{D}=(\mathcal{G}_x, \mathcal{Y}, \mathcal{G}_y)$, SRIN training epochs $T_1$, DRAM training epochs $T_2$, Model training iterations $T$\\
    % \textbf{Parameter}: Optional list of parameters\\
    \textbf{Output}: Prediction result $\hat{\mathcal{Y}}$
    \begin{algorithmic}[1]
        \STATE Convert SMILES sequences into graph structures;
        \STATE Initialize training parameters;
        \FOR{$t=1$ to $T$}{
        \FOR{$t_1=1$ to $T_1$}{
        \STATE Compute substructure embeddings $\mathcal{Z}(\mathcal{G}^s)$ by Eq (\ref{irr});
        \STATE Obtain interaction-optimized substructure representations $\mathcal{Z}(\mathcal{G}'^s)$ based on Eqs (\ref{eq:interpro}) and (\ref{eq:gen});
        }
        \ENDFOR
        \FOR{$t_2=1$ to $T_2$}{
        \STATE Generate reconstructed substructures $\mathcal{Z}(\mathcal{G}^{news})$ via $e_{ik}$;
        \STATE Obtain aligned core substructure pairs according to Eqs (\ref{align}) and (\ref{if});
        \STATE Optimize S-GIB and generate refined substructure pairs with Eqs (\ref{klxc}) and (\ref{loss});
        \IF{$\mathcal{L}$ in Eq (\ref{lossnum}) converges}
        \STATE \textbf{Return} the S-GIB-optimized representation $\mathcal{H}$;
        \ELSE
        \STATE $t_2 \leftarrow t_2+1$;
        \ENDIF
        }
        \ENDFOR
        }
        \ENDFOR
        \STATE Output the prediction result $\hat{\mathcal{Y}}$, computed from $\mathcal{H}$.
    \end{algorithmic}
\end{algorithm}
\begin{table}[htpb]
\small
    \centering
    \renewcommand{\arraystretch}{1}
    \caption{Data statistics for different datasets.}
    \setlength{\tabcolsep}{0.6mm}{
    \begin{tabular}{c|c|ccccccc}
    \hline
        \multicolumn{2}{c|}{Dataset} & \#$G_x$ & \#$G_y$ & \#Pairs & \#Tra & \#Val & \#Tes  & Task\\
        \hline
        \multirow{3}{*}{Chromophore} & Absorption & 6416 & 725 & 17276 & 10366 & 3455 & 3455 & MI \\	
        &  Emission & 6412 & 1021 & 18141 & 10885 & 3628 & 3628 & MI\\
        & Lifetime & 2755 & 247 & 6960 & 4176 & 1392 & 1392 & MI\\
        \hline
        \multicolumn{2}{c|}{MNSol} & 372 & 86 & 2275 & 1365 & 455 & 455 & MI\\
        \multicolumn{2}{c|}{FreeSolv} & 560 & 1 & 560 & 336 & 112 & 112 & MI\\
        \multicolumn{2}{c|}{CompSol} & 442 & 259 & 3548 & 2130 & 709 & 709 & MI\\
        \multicolumn{2}{c|}{Abraham} & 1038 & 122 & 6091 & 3655 & 1218 & 1218 & MI\\
        \multicolumn{2}{c|}{CombiSolv} & 1495 & 326 & 10145 & 6087 & 2029 & 2029 & MI\\
        \hline
        \multicolumn{2}{c|}{ZhangDDI} & 544 & 544 & 40255 & 48306 & 16102 & 16102 & DDI\\
        \multicolumn{2}{c|}{HetionteDDI} & 696 & 696 & 6410 & 7692 & 2564 & 2564 & DDI \\
        \multicolumn{2}{c|}{DrugBankDDI} & 2045 & 2045 & 291822 & 350186 & 116729 & 116729 & DDI \\
        \hline
    \end{tabular} 
    }
    \label{dataset}
\end{table}

\begin{table*}[t]
\caption{The performance of ReAlignFit and comparative methods in MI prediction, with the best results highlighted in \colorbox{c1!80}{\textbf{bold}} and the second results highlighted in \colorbox{c1!40}{text}.}
\small
    \centering
    \renewcommand{\arraystretch}{1}
    \setlength{\tabcolsep}{2mm}{
    \begin{tabular}{lcccccccc}
    \hline
        \multirow{2}{*}{Model} & \multicolumn{3}{c}{Chromophore} & \multirow{2}{*}{MNSol} & \multirow{2}{*}{FreeSolv} & \multirow{2}{*}{CompSol} & \multirow{2}{*}{Abraham} & \multirow{2}{*}{CombiSolv}\\
        \cline{2-4}
         & Absorption & Emission & Lifetime &  &  &  &  &  \\
          \hline
          \multicolumn{9}{c}{\textbf{No Representational Alignment}}\\
          \hline
        GCN(ICLR'17) & 25.75$\pm$1.48 & 31.87$\pm$1.70 & 0.866$\pm$0.015 & 0.675$\pm$0.021 & 1.192$\pm$0.042 & 0.389$\pm$0.009 & 0.738$\pm$0.041  & 0.672$\pm$0.022\\
        GAT(ICLR'18) & 26.19$\pm$1.44 & 30.90$\pm$1.01 & 0.859$\pm$0.016 & 0.731$\pm$0.007 & 1.280$\pm$0.049 & 0.387$\pm$0.010 & 0.798$\pm$0.038 & 0.662$\pm$0.021\\
        MPNN(ICML'17) & 24.43$\pm$1.55 & 30.17$\pm$0.99 & 0.802$\pm$0.024 & 0.682$\pm$0.017 & 1.159$\pm$0.032 & 0.359$\pm$0.011  & 0.601$\pm$0.035 & 0.568$\pm$0.005\\
       GIN(ICLR'19) & 24.92$\pm$1.67 & 32.31$\pm$0.26 & 0.829$\pm$0.027 & 0.669$\pm$0.017 & 1.015$\pm$0.041 & 0.331$\pm$0.016 & 0.648$\pm$0.024 & 0.595$\pm$0.014\\
       CIGIN(AAAI'20) & 19.32$\pm$0.35 & 25.09$\pm$0.32 & 0.804$\pm$0.010 & 0.607$\pm$0.024 & 0.905$\pm$0.014 & 0.308$\pm$0.018 & 0.411$\pm$0.008 & 0.451$\pm$0.009\\
        \hline
        \multicolumn{9}{c}{\textbf{Representational Alignment by Attention-based Inductive Bias}}\\
        \hline
       CMRL(KDD'23) & 17.93$\pm$0.31 & 24.30$\pm$0.22 & 0.776$\pm$0.007 & 0.551$\pm$0.017 & 0.815$\pm$0.046 & 0.255$\pm$0.011 & 0.374$\pm$0.011 & 0.421$\pm$0.008\\
       CGIB(ICML'23) & 18.11$\pm$0.20 & 23.90$\pm$0.35 & \cellcolor{c1!40}{0.771$\pm$0.005} & \cellcolor{c1!80}\textbf{0.538$\pm$0.007} & 0.852$\pm$0.022 & 0.276$\pm$0.017 & 0.390$\pm$0.006 & 0.422$\pm$0.005\\
       MMGNN(IJCAI'24)& 18.65$\pm$0.34 & 25.33$\pm$0.43 & 0.801$\pm$0.007 & 0.546$\pm$0.011 & 0.902$\pm$0.026 & 0.267$\pm$0.012 & 0.385$\pm$0.008 & \cellcolor{c1!80}\textbf{0.303$\pm$0.033} \\
       ISE(ICLR'25) & 17.81$\pm$0.37 & 24.66$\pm$0.42 & 0.773$\pm$0.026 & 0.607$\pm$0.028 & 0.825$\pm$0.039 & 0.268$\pm$0.013 & \cellcolor{c1!80}\textbf{0.369$\pm$0.014} & 0.400$\pm$0.010\\
        \hline
        ReAlignFit & \cellcolor{c1!80}\textbf{16.25$\pm$0.21} & \cellcolor{c1!80}\textbf{21.99$\pm$0.29} & \cellcolor{c1!80}\textbf{0.756$\pm$0.005} & 0.540$\pm$0.009 & \cellcolor{c1!40}{0.798$\pm$0.030} & 0.259$\pm$0.012 & \cellcolor{c1!40}\textbf{0.370$\pm$0.008} & 0.367$\pm$0.006\\
        ReAlignFit$_\text{GCN}$ & \cellcolor{c1!40}{17.23$\pm$0.27} & \cellcolor{c1!40}{23.35$\pm$0.29} & 0.771$\pm$0.007 & \cellcolor{c1!40}{0.539$\pm$0.012} & \cellcolor{c1!80}\textbf{0.796$\pm$0.035} & \cellcolor{c1!40}{0.257$\pm$0.016} & 0.375$\pm$0.012 & \cellcolor{c1!40} \textbf{0.316$\pm$0.006}\\
        ReAlignFit$_\text{GAT}$ & 17.55$\pm$0.23 & 23.98$\pm$0.36 & 0.776$\pm$0.007 & 0.543$\pm$0.021 & 0.806$\pm$0.036 & \cellcolor{c1!80}\textbf{0.254$\pm$0.012} & 0.381$\pm$0.013 & 0.421$\pm$0.009\\
        ReAlignFit$_\text{GIN}$ & 17.92$\pm$0.46 & 24.10$\pm$0.34 & 0.772$\pm$0.007 & 0.552$\pm$0.037 & 0.803$\pm$0.025 & 0.267$\pm$0.021 & 0.386$\pm$0.017 & 0.318$\pm$0.008\\
        \hline
    \end{tabular}
    } 
    % The results of comparison methods are from~\cite{lee2023cgibICML,lee2023shiftKDD}.}
    \label{tab:mi}
\end{table*}

\section{Experiments}
\label{Experiments}
In this section, we aim to answer the following research questions by analyzing the relevant experiments: \textbf{RQ1}: How does ReAlignFit perform in MRL, and whether it is susceptible to backbone? \textbf{RQ2}: Can ReAlignFit improve the stability of MRL in distribution-shifted data? \textbf{RQ3}: Can the results of ReAlignFit be visually supported?

\section{Experimental Setup}
\label{Experimental Setup}

\subsection{Datasets}
Following the related research \cite{lee2023cgibICML,lee2023shiftKDD,boulougouri2024molecularMNI,du2024mmgnnIJCAI}, we conduct extensive Molecular Interaction (MI) prediction and Drug-Drug Interaction (DDI) prediction experiments on nine datasets, as detailed in Table \ref{dataset}. Chromophore \cite{chromophore}, MNSol \cite{mnsol}, FreeSolv \cite{freesolv}, CompSol \cite{comsol}, Abraham \cite{abraham}, and CombiSolv \cite{combisolv} are chromophore datasets used to describe the free energy of solutes in solvents. These six datasets are widely used for MI prediction \cite{lee2023cgibICML,lee2023shiftKDD,du2024mmgnnIJCAI}. ZhangDDI \cite{zhangddi}, HetionetDDI \cite{hddi}, and DrugBankDDI (paid version) \cite{drugbankddi} are commonly utilized for DDI prediction \cite{zhang2024heterogeneousIJCAI}. Following the setup of related work in MRL \cite{lee2023cgibICML,lee2023shiftKDD}, we divide the datasets into training, validation, and test sets with the ratio of 6:2:2. For DDI datasets that contain only positive examples, we generated negative samples using rule matching and scaffold clustering methods, respectively. To better simulate the real-world data distribution, we construct in-distribution data (Original) with the same distribution as the original data and two different distribution-shifted data (P1 and P2) for DDI prediction, which were used for ReAlignFit learning.

\begin{itemize}
    \item \textbf{In-Distribution Data (Original)}: Negative samples are generated from the original data based on rule matching and the dataset is then randomly divided into training, validation, and test sets.
    \item \textbf{Rule-based Partitioning (P1)}: We generated negative samples for original data based on rule matching and partitioned them by ID to ensure that at least one drug doesn't repeat in training, validation, and test sets. P1 simulates application scenarios of discovering unknown interactions in existing molecules, such as drug repurposing.
    \item \textbf{Scaffold-based Partitioning (P2)}: We generated negative samples for the original data based on scaffold clustering. We used METIS \cite{karypis1998multilevelMETIS} to iteratively partition the drug interaction graph, ensuring that molecules in training, validation, and test sets are entirely distinct. P2 simulates scenarios of discovering interactions between previously unknown molecules, such as drug discovery.
\end{itemize}

\subsection{Evaluation Metrics}
In this study, we treat MI prediction and DDI prediction as regression and classification tasks, respectively. Root Mean Square Error (RMSE) and Mean Absolute Error (MAE) are commonly used metrics in regression prediction. Considering the high positive correlation between RMSE and MAE, we employ RMSE for performance evaluation in MI prediction. We assess model performance for DDI prediction based on classification evaluation metrics, including Area Under the Receiver Operating Characteristic Curve (AUROC), Accuracy (ACC), F1-score (F1), Precision (Pre), and Area Under the Precision-Recall Curve (AUPR).

\subsection{Training Details}
In all experiments, we employ a three-layer MPNN as the encoder to extract molecular feature representations and train ReAlignFit based on two NVIDIA GeForce RTX 4090 24G GPUs. We optimize the parameters using the Adam optimizer and train the model for 50 epochs. The batch size is 128, and the dropout rate is 0.1. For other hyperparameters, we choose from a specific range: learning rate $lr \in \{0.01,0.005,0.001,0.0005,0.0001\}$, both $\alpha$ and $\beta$ are selected from $\{0.5,0.3,0.1,0.01,0.001\}$, and the iterations is searched from $\{1,3,5,10,15\}$. The code of ReAlignFit will be publicly available after the paper is accepted.

\subsection{Baseline Models}
We compare ReAlignFit with four backbone models four backbone models (GCN~\cite{gcn}, GAT~\cite{gat}, MPNN~\cite{mpnn}, and GIN~\cite{gin}), three molecular representation-based models (CIGIN~\cite{cigin}, MIRACLE~\cite{miracle}, and MMGNN~\cite{du2024mmgnnIJCAI}), and five substructure representation-based models (SSI-DDI~\cite{ssiddi}, SA-DDI~\cite{saddi}, DSN-DDI~\cite{dsn}, CMRL~\cite{lee2023shiftKDD}, CGIB~\cite{lee2023cgibICML}, and ISE~\cite{zhang2025iterative}) on MI and DDI tasks.

\begin{table*}[t]
\caption{The performance of ReAlignFit and comparative methods in DDI prediction, with the best results highlighted in \colorbox{c1!80}{\textbf{bold}} and the second results highlighted in \colorbox{c1!40}{text}.} 
    \label{tab:ddi}
\small
    \centering
    \renewcommand{\arraystretch}{1}
    \setlength{\tabcolsep}{1.3mm}{
    \begin{tabular}{lccccc|ccccc|ccccc}
    \hline
        \multirow{2}{*}{Model} & \multicolumn{5}{c|}{ZhangDDI} & \multicolumn{5}{c|}{HetionteDDI} & \multicolumn{5}{c}{DrugBankDDI}\\
        \cline{2-16}
         & ACC & AUROC & F1 & Pre & AUPR & ACC & AUROC & F1 & Pre & AUPR & ACC & AUROC & F1 & Pre & AUPR  \\ \hline
         \multicolumn{16}{c}{\textbf{No Representational Alignment}}\\
          \hline
        GCN(ICLR'17) & 83.31 & 91.64 & 82.91 & 77.67 & 91.33 & 87.62 & 92.91 & 90.07 & 84.66 & 91.26 & 84.02 & 90.96 & 82.87 & 80.04 & 88.36  \\
        GAT(ICLR'18) & 84.14 & 92.10 & 83.42 & 78.45 & 91.21 & 87.94 & 94.35 & 89.78 & 83.52 & 92.83 & 84.33 & 91.26 & 83.34 & 80.58 & 89.13  \\
        MPNN(ICML'17) & 84.56 & 92.34 & 83.32 & 78.15 & 91.46 & 88.73 & 95.13 & 89.52 & 83.66 & 93.21 & 87.30 & 95.90 & 88.53 & 80.73 & 95.08  \\
        GIN(ICLR'19) & 85.59 & 93.16 & 85.07 & 79.82 & 92.11 & 87.37 & 93.01 & 87.91 & 84.05 & 91.23 & 85.21 & 92.58 & 85.88 & 83.57 & 90.92  \\
        MIRACLE(WWW'21) & 84.90 & 93.05 & 83.94 & 77.81 & 91.25 & 87.01 & 92.88 & 87.34 & 83.91 & 90.88 & 84.98 & 92.17 & 85.43 & 83.22 & 90.28  \\
        CIGIN(AAAI'20) & 85.54 & 93.28 & 84.36 & 80.23 & 91.89 & 90.87 & 92.77 & 90.33 & 88.98 & 94.62 & 91.02 & 93.62 & 91.18 & 89.35 & 94.88  \\
        \hline
         \multicolumn{16}{c}{\textbf{Representational Alignment by Attention-based Inductive Bias}}\\
          \hline
        SSI-DDI(BIB'21) & 85.35 & 93.14 & 81.96 & 78.97 & 92.09 & 87.95 & 93.83 & 88.56 & 84.28 & 91.60 & 84.16 & 91.23 & 84.64 & 82.13 & 89.47  \\
        DSN-DDI(BIB'23) & 86.65 & 91.13 & 87.86 & 84.33 & 86.42 & 91.25 & 95.31 & 91.59 & 89.52 & 94.66 & 90.49 & 96.15 & 90.58 & 89.76 & 95.27  \\
        SA-DDI(CS'22) & 77.31 & 50.26 & 45.06 & 52.54 & 29.97 & 91.00 & \cellcolor{c1!40}{95.89} & 91.26 & 88.39 & 94.96 & 92.60 & 95.33 & 92.80 & 90.38 & 96.62  \\
        CMRL(KDD'23) & 86.32 & 93.73 & 87.68 & 84.01 & 91.56 & 91.25 & 93.18 & 91.53 & 91.23 & 95.24 & 92.26 & 95.05 & 92.43 & 90.30 & 95.96  \\
        CGIB(ICML'23) & 86.36 & 93.78 & 87.24 & 83.91 & 91.88 & 91.08 & 93.26 & 91.35 & 91.55 & 94.89 & 92.37 & 94.98 & 92.03 & 90.65 & 96.11  \\ 
        MMGNN(IJCAI'24) & 85.33 & 93.23 & 86.45 & 84.36 & 90.89 & 90.67 & 94.34 & 90.75 & 92.22 & \cellcolor{c1!40}{95.28} & 93.61 & 95.12 & 92.38 & 91.22 & 95.39\\
        ISE(ICLR'25) & \cellcolor{c1!40}{88.45} & 94.25 & 88.65 & 85.32 & 91.62 & 90.82 & 94.36 & 90.51 & 93.01 & 94.96 & 94.56 & 96.35 & 93.18 & 92.33 & 95.57\\
        \hline
        ReAlignFit & \cellcolor{c1!80}\textbf{89.43} & \cellcolor{c1!40}{95.68} & \cellcolor{c1!80}\textbf{90.88} & \cellcolor{c1!80}\textbf{87.68} & \cellcolor{c1!80}\textbf{93.36} & \cellcolor{c1!80}\textbf{92.39} & \cellcolor{c1!80}\textbf{97.17} & \cellcolor{c1!80}\textbf{92.62} & \cellcolor{c1!80}\textbf{95.48} & \cellcolor{c1!80}\textbf{96.05} & \cellcolor{c1!40}{95.53} & 96.31 & 94.59 & \cellcolor{c1!80}\textbf{94.38} & \cellcolor{c1!40}{97.05} \\ 
        ReAlignFit$_\text{GCN}$ & 87.62 & \cellcolor{c1!80}\textbf{95.69} & 90.34 & \cellcolor{c1!40}{87.54} & \cellcolor{c1!40}{92.81} & 91.16 & 94.22 & 91.55 & 92.29 & 95.21 & \cellcolor{c1!80}\textbf{95.62} & 97.95 & \cellcolor{c1!80}\textbf{95.62} & \cellcolor{c1!40}{94.37} & \cellcolor{c1!80}\textbf{97.09} \\
        ReAlignFit$_\text{GAT}$ & 88.35 & 94.22 & \cellcolor{c1!40}{90.41} & 86.92 & 91.76 & 91.66 & 94.56 & 91.76 & 91.97 & 95.03 & 95.35 & \cellcolor{c1!80}\textbf{98.42} & 93.87 & 93.67 & 96.26 \\
        ReAlignFit$_\text{GIN}$ & 85.23 & 94.35 & 89.34 & 86.89 & 90.67 & \cellcolor{c1!40}{92.12} & 92.95 & \cellcolor{c1!40}{91.91} & \cellcolor{c1!40}{93.04} & 95.23 & 93.33 & \cellcolor{c1!40}{98.09} & \cellcolor{c1!40}{95.01} & 91.58 & 95.82\\
        \hline
    \end{tabular}
    }
\end{table*}

\subsection{Overall Performance (\textbf{RQ1})}
The MI prediction and DDI prediction results are reported in Table \ref{tab:mi} and Table \ref{tab:ddi}, respectively.

\textbf{Comprehensive Analysis}: ReAlignFit achieves the best overall performance in MI prediction and DDI prediction. Particularly, ReAlignFit achieves an average improvement of 2.7\% in ACC and 3.0\% in AUROC in three DDI datasets. Additionally, ReAlignFit achieves the best prediction performance in multiple MI datasets, reducing RMSE by 1.83 and 2.38 of Absorption and Emission attributes on the Chromophore dataset, respectively. These result suggests that incorporating domain knowledge into MRL and substructure representational alignment contributes to improving the model's predictive performance. 

\textbf{Importance Analysis of Representational Alignment}: The prediction performance of methods considering representational alignment (CGIB, MMGNN, ReAlignFit, etc.) is significantly better than those that ignore it. For both DDI and MI prediction, the model performance improvement by representational alignment is noticeable. This result indicates representational alignment is a critical factor in molecular relationships. However, it is worth noting that representational alignment methods by attention-based inductive bias achieve an average improvement of less than 5\% on the Chromophore dataset, whereas ReAlignFit exceeds 8.5\%. This observation is consistent with the nature of the Induced Fit, emphasizes dynamic conformational changes of substructures.

\textbf{ReAlignFit Compatibility Analysis}: By substituting different GNN models within ReAlignFit, we observe minor performance fluctuations. ReAlignFit, combined with different GNN backbones, consistently achieves the best or second-best predictive performance in all evaluation metrics for FreeSolv, CompSol, ZhangDDI and DrugBankDDI datasets. This result indicates that ReAlignFit is less affected by the backbone, demonstrating flexibility in integration with general graph models. Furthermore, this highlights ReAlignFit's reliable applicability at the model level.

\subsection{Stability Analysis (\textbf{RQ2})}
\begin{figure}[t]
\centering
\includegraphics[width=\columnwidth]{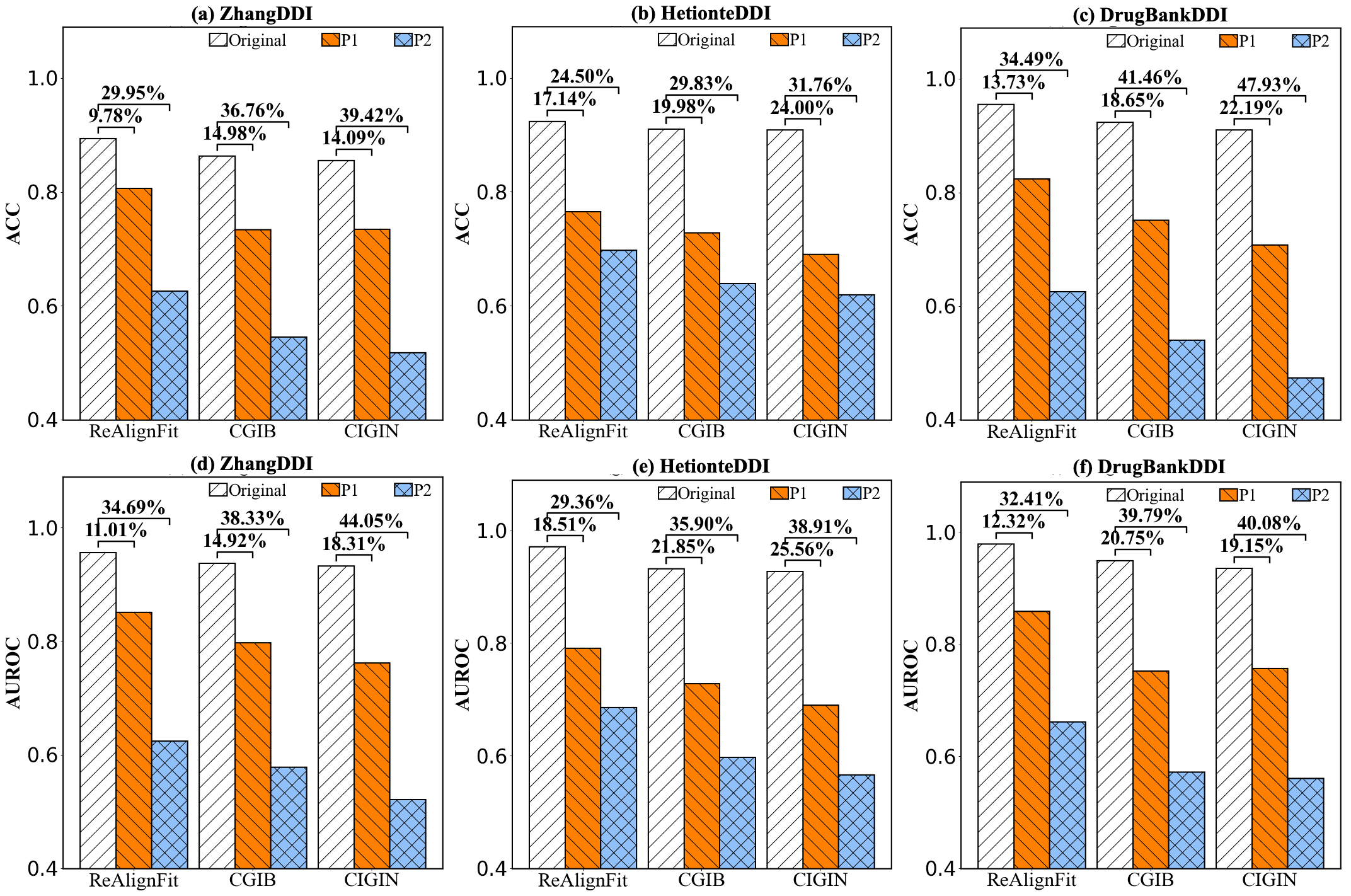}
\caption{The performance and RPD of ReAlignFit, CGIB and CIGIN in different data distributions.}
\label{stableresults}
\end{figure}

In this section, we analyze the stability of model prediction performance in distribution-shifted data. 
To evaluate the model's performance in different data distributions, we quantify model stability by calculating the Relative Performance Degradation (RPD) for AUROC, ACC, F1, Pre, and AUPR. RPD is defined as follows:
\begin{equation}
    RPD_{M}=\frac{|Eva_{M}^{P_i}-Eva_{M}^{Ori}|}{Eva_{M}^{Ori}}\times 100\%
\end{equation}
where $Eva$ represents the evaluation metrics. $Eva_{M}^{P}$ and $Eva_{M}^{Ori}$ denote the prediction performance in the original and drift distributions, respectively. $M \in {AUROC, ACC, F1, Pre, AUPR}$ and ${P_i}\in {P1,P2}$.
The experimental results are shown in Fig. \ref{stableresults}. The \textbf{values} in Fig. \ref{stableresults} represent RPD. 
Among these, P1 exhibits the most minor distributional differences from the original data, followed by P2. 

\begin{figure*}[t]
\centering
\includegraphics[width=\textwidth]{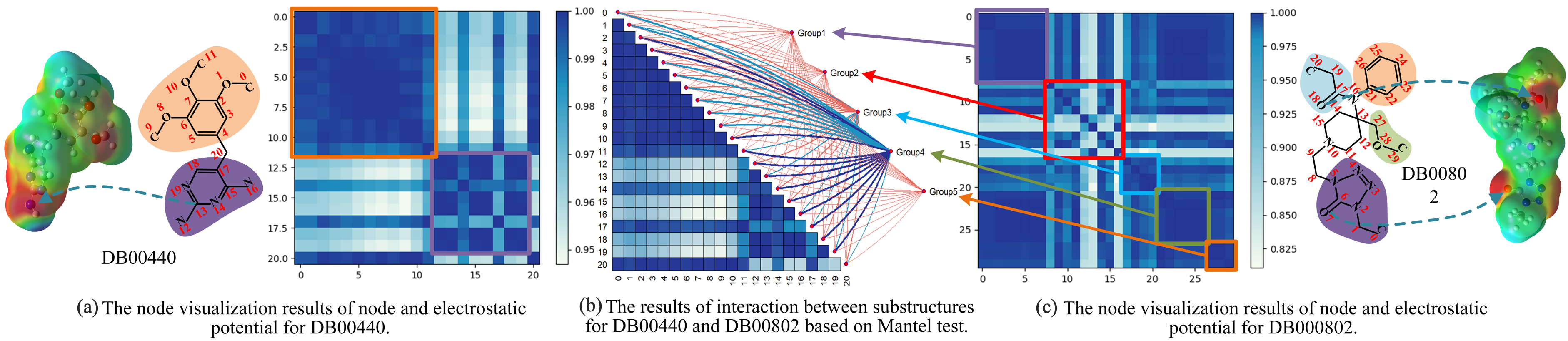}
\caption{The visualization of node features and interaction strengths between substructures.}
\label{vis}
\end{figure*}

\textbf{Stability Analysis}: Experimental results show that data distribution variations significantly impact the predictive performance of models. 
ReAlignFit maintains the best stability in different data distributions compared to methods that do not consider representational alignment. Compared to CGIB, ReAlignFit improved ACC prediction stability by at least 4.5\% and 6\% in two data distribution scenarios. ReAlignFit exhibits the most minor performance degradation and achieves the highest stability in varying data distributions. ReAlignFit's performance degradation ranging from 6\% to 18\% in the P1 scenario, with certain evaluation metrics showing only around 25\% decline in P2. In contrast, CGIB and CIGIN experience performance fluctuations exceeding 35\% in most metrics on P2. These findings show that stability is the critical factor in the success of ReAlignFit, primarily attributed to DRAM, which identifies chemically functionally compatible core substructure pairs through representational alignment. In summary, representational alignment improves model stability and maintains strong predictive performance in distribution-shifted datasets. This directly confirms the role of representational alignment in improving the stability.

\textbf{Predictive Performance Analysis:}
With increasing distributional differences, all models exhibit varying degrees of performance degradation, highlighting that data distribution shifts remain a significant challenge for MRL. 
Encouragingly, ReAlignFit consistently outperforms other comparison models in DDI prediction in the P2 data scenario. 
Moreover, ReAlignFit outperforms the second-best method in three datasets on the P1 data scenario by 5.07\%, 3.67\%, 5.19\% in ACC, and 3.32\%, 6.05\%, 1.87\% in AUROC. These results provide the possibility of applying ReAlignFit in molecular science. By comprehensively analyzing the experimental results of ReAlignFit in distribution drift scenarios, we can see that ReAlignFit exhibits better prediction performance and stability in distribution-shifted data. We attribute the results to two main reasons: substructure representational alignment identifies substructure pairs that contain stable causal information.
On the other hand, ReAlignFit mitigates the impact of confounding substructures on molecular representations by weakening the association of confounding substructures with causal substructures.

\subsection{Visualization Analysis (RQ3)}
To validate ReAlignFit's ability to capture core substructures in MRL, we visualize the node features and the interactions between substructures, as shown in Fig. \ref{vis}.

In the node feature representations of drug molecules, the substructure information captured by ReAlignFit closely aligns with the actual electrostatic potential of the molecules and emphasizes the interactions between atoms within molecules. As shown in Figs. \ref{vis}\textcolor{blue}{(a), (c)}, ReAlignFit demonstrates strong recognition capabilities for multiple complex functional groups within molecules. 
The Mantel test~\cite{manteltest} results shown in the Fig. \ref{vis}\textcolor{blue}{(b)} illustrate the interaction strength between different substructures. Darker and thicker lines represent stronger interactions between substructures.
Specifically, the results indicate the high interaction strength between the functional group (purple region) in DB00440 and Group4 in DB00802. 
These findings are consistent with the Induced Fit theory which posits that chemical reactions occur between core substructures.
In summary, ReAlignFit not only captures substructure information within molecules but also highlights the crucial role of this substructure information in determining the occurrence of chemical reactions. The experimental results provide strong evidence for ReAlignFit to capture core substructures in MRL.

\section{Limitations and Ethical Considerations}
Although ReAlignFit improves the stability of MRL under distribution shift through a chemically induced fit mechanism, its effectiveness in dynamically reconstructing and aligning substructures depends on the comprehensiveness of the substructure partitioning. Furthermore, the stability of ReAlignFit is primarily validated under rule-based and scaffold-based shift settings. Its generalization capability under more complex real-world open-environment shifts (such as cross-platform variation, cross-experimental conditions, and cross-modal distribution gaps) remains to be further investigated. This study uses only publicly datasets, involving no personal information.

\section{Conclusion}
In this paper, we propose the the Representational Alignment with Chemical Induced Fit to improve the stability of MRL. ReAlignFit dynamically aligns representation by introducing Induced Fit-based inductive bias. ReAlignFit simulates dynamic conformational changes during Induced Fit by a self-supervised approach. Additionally, it integrates S-GIB to align core substructure pairs with potential compatibility in chemical functions. Experimental results demonstrate that ReAlignFit achieves the best predictive performance and significantly improves the predictive stability in distribution-shifted data.

\section*{Acknowledgement}
This work was supported by the National Natural Science Foundation of China (No. 62472332 and No. 62276196), the Hainan Provincial Natural Science Foundation of China (No. 526MS0269), the Basic Project for Universities from the Educational Department of Liaoning Province (No. LJ242511258004), and the 111 Project (No. D23006).

\bibliographystyle{ACM-Reference-Format}
\bibliography{sample-base}

@String{Computing = "Computing" }

@String{Computer = "{IEEE} Computer" }

@String{Springer = "Springer-Verlag" }

@ArtifactSoftware{R,
    title = {R: A Language and Environment for Statistical Computing},
    author = {{R Core Team}},
    organization = {R Foundation for Statistical Computing},
    address = {Vienna, Austria},
    year = {2019},
    url = {https://www.R-project.org/},
}

@article{zhao2025evidential,
  title={Evidential deep learning-based drug-target interaction prediction},
  author={Zhao, Yanpeng and Xing, Yuting and Zhang, Yixin and Wang, Yifei and Wan, Mengxuan and Yi, Duoyun and Wu, Chengkun and Li, Shangze and Xu, Huiyan and Zhang, Hongyang and others},
  journal={Nature communications},
  volume={16},
  number={1},
  pages={6915},
  year={2025},
  publisher={Nature Publishing Group UK London}
}

@article{wang2025image,
  title={Image-based generation for molecule design with SketchMol},
  author={Wang, Zixu and Chen, Yangyang and Ma, Pengsen and Yu, Zhou and Wang, Jianmin and Liu, Yuansheng and Ye, Xiucai and Sakurai, Tetsuya and Zeng, Xiangxiang},
  journal={Nature Machine Intelligence},
  volume={7},
  number={2},
  pages={244--255},
  year={2025},
  publisher={Nature Publishing Group UK London}
}

@article{lu2025dtiam,
  title={DTIAM: a unified framework for predicting drug-target interactions, binding affinities and drug mechanisms},
  author={Lu, Zhangli and Song, Guoqiang and Zhu, Huimin and Lei, Chuqi and Sun, Xinliang and Wang, Kaili and Qin, Libo and Chen, Yafei and Tang, Jing and Li, Min},
  journal={Nature Communications},
  volume={16},
  number={1},
  pages={2548},
  year={2025},
  publisher={Nature Publishing Group UK London}
}

@inproceedings{wang2024kdd,
  title="{Advancing Molecule Invariant Representation via Privileged Substructure Identification}",
  author={Wang, Ruijia and Dai, Haoran and Yang, Cheng and Song, Le and Shi, Chuan},
  booktitle={Proceedings of the 30th ACM SIGKDD Conference on Knowledge Discovery and Data Mining},
  pages={3188--3199},
  year={2024}
}

@InProceedings{lee2023cgibICML,
  title = 	 "{Conditional Graph Information Bottleneck for Molecular Relational Learning}",
  author =       {Lee, Namkyeong and Hyun, Dongmin and Na, Gyoung S. and Kim, Sungwon and Lee, Junseok and Park, Chanyoung},
  booktitle = 	 {Proceedings of the 40th International Conference on Machine Learning},
  pages = 	 {18852--18871},
  year = 	 {2023},
  volume = 	 {202}
}

@inproceedings{lee2023shiftKDD,
  title="{Shift-Robust Molecular Relational Learning with Causal Substructure}",
  author={Lee, Namkyeong and Yoon, Kanghoon and Na, Gyoung S and Kim, Sein and Park, Chanyoung},
  booktitle={Proceedings of the 29th ACM SIGKDD Conference on Knowledge Discovery and Data Mining},
  pages={1200--1212},
  year={2023}
}

@article{boulougouri2024molecularMNI,
  title="{Molecular Set Representation Learning}",
  author={Boulougouri, Maria and Vandergheynst, Pierre and Probst, Daniel},
  journal={Nature Machine Intelligence},
  pages={754--763},
  volume={6},
  year={2024},
  publisher={Nature Publishing Group UK London}
}

@inproceedings{seo2024selfKDD,
  title="{Self-Explainable Temporal Graph Networks based on Graph Information Bottleneck}",
  author={Seo, Sangwoo and Kim, Sungwon and Jung, Jihyeong and Lee, Yoonho and Park, Chanyoung},
  booktitle={Proceedings of the 30th ACM SIGKDD Conference on Knowledge Discovery and Data Mining},
  year={2024}
}

@inproceedings{yang2023molerecwww,
  title="{Molerec: Combinatorial Drug Recommendation with Substructure-Aware Molecular Representation Learning}",
  author={Yang, Nianzu and Zeng, Kaipeng and Wu, Qitian and Yan, Junchi},
  booktitle={Proceedings of the 32nd ACM on Web Conference},
  pages={4075--4085},
  year={2023}
}

@inproceedings{su2024dualAAAI,
  title="{Dual-Channel Learning Framework for Drug-Drug Interaction Prediction via Relation-Aware Heterogeneous Graph Transformer}",
  author={Su, Xiaorui and Hu, Pengwei and You, Zhu-Hong and Philip, S Yu and Hu, Lun},
  booktitle={Proceedings of the 38th AAAI Conference on Artificial Intelligence},
  volume={38},
  number={1},
  pages={249--256},
  year={2024}
}

@inproceedings{zhang2024heterogeneousIJCAI,
  title="{Heterogeneous Causal Metapath Graph Neural Network for Gene-Microbe-Disease Association Prediction}",
  author={Zhang, Kexin and Huang, Feng and Liu, Luotao and Xiong, Zhankun and Zhang, Hongyu and Quan, Yuan and Zhang, Wen},
  booktitle={Proceedings of the 33rd International Joint Conference on Artificial Intelligence},
  year={2024}
}

@article{saddi,
  title="{Learning Size-Adaptive Molecular Substructures for Explainable Drug--Drug Interaction Prediction by Substructure-Aware Graph Neural Network}",
  author={Yang, Ziduo and Zhong, Weihe and Lv, Qiujie and Chen, Calvin Yu-Chian},
  journal={Chemical Science},
  volume={13},
  number={29},
  pages={8693--8703},
  year={2022},
  publisher={Royal Society of Chemistry}
}

@article{ssiddi,
  title="{SSI--DDI: Substructure--Substructure Interactions for Drug--Drug Interaction Prediction}",
  author={Nyamabo, Arnold K and Yu, Hui and Shi, Jian-Yu},
  journal={Briefings in Bioinformatics},
  volume={22},
  number={6},
  pages={bbab133},
  year={2021},
  publisher={Oxford University Press}
}

@article{dsn,
  title="{DSN-DDI: An Accurate and Generalized Framework for Drug--Drug Interaction Prediction by Dual-View Representation Learning}",
  author={Li, Zimeng and Zhu, Shichao and Shao, Bin and Zeng, Xiangxiang and Wang, Tong and Liu, Tie-Yan},
  journal={Briefings in Bioinformatics},
  volume={24},
  number={1},
  pages={bbac597},
  year={2023},
  publisher={Oxford University Press}
}

@inproceedings{cigin,
  title="{Chemically Interpretable Graph Interaction Network for Prediction of Pharmacokinetic Properties of Drug-Like Molecules}",
  author={Pathak, Yashaswi and Laghuvarapu, Siddhartha and Mehta, Sarvesh and Priyakumar, U Deva},
  booktitle={Proceedings of the 34th AAAI Conference on Artificial Intelligence},
  volume={34},
  number={01},
  pages={873--880},
  year={2020}
}

@article{nc,
  title="{A Variational Expectation-Maximization Framework for Balanced Multi-Scale Learning of Protein and Drug Interactions}",
  author={Rao, Jiahua and Xie, Jiancong and Yuan, Qianmu and Liu, Deqin and Wang, Zhen and Lu, Yutong and Zheng, Shuangjia and Yang, Yuedong},
  journal={Nature Communications},
  volume={15},
  number={1},
  pages={4476},
  year={2024},
  publisher={Nature Publishing Group UK London}
}

@inproceedings{karypis1998multilevelMETIS,
  title="{Multilevel Algorithms for Multi-Constraint Graph Partitioning}",
  author={Karypis, George and Kumar, Vipin},
  booktitle={Proceedings of the ACM/IEEE Conference on Supercomputing},
  pages={28--28},
  year={1998}
}

@inproceedings{miracle,
  title="{Multi-View Graph Contrastive Representation Learning for Drug-Drug Interaction Prediction}",
  author={Wang, Yingheng and Min, Yaosen and Chen, Xin and Wu, Ji},
  booktitle={Proceedings of the 30th ACM on Web Conference},
  pages={2921--2933},
  year={2021}
}

@inproceedings{gcn,
  title="{Semi-Supervised Classification with Graph Convolutional Networks}",
  author={Kipf, Thomas N and Welling, Max},
  booktitle={Proceedings of the 5th International Conference on Learning Representations},
  year={2017}
}

@inproceedings{mpnn,
author = {Gilmer, Justin and Schoenholz, Samuel S. and Riley, Patrick F. and Vinyals, Oriol and Dahl, George E.},
title = "{Neural Message Passing for Quantum Chemistry}",
year = {2017},
booktitle = {Proceedings of the 34th International Conference on Machine Learning},
pages = {1263–1272}
}

@inproceedings{gin,
  title="{How Powerful are Graph Neural Networks?}",
  author={Xu, Keyulu and Hu, Weihua and Leskovec, Jure and Jegelka, Stefanie},
  booktitle={Proceedings of the 7th International Conference on Learning Representations},
  year={2019}
}

@inproceedings{gat,
  title="{Graph Attention Networks}",
  author={Veli{\v{c}}kovi{\'c}, Petar and Cucurull, Guillem and Casanova, Arantxa and Romero, Adriana and Lio, Pietro and Bengio, Yoshua},
  booktitle={Proceedings of the 6th International Conference on Learning Representations},
  year={2018}
}

@ARTICLE{tkdeGIBB,
  author={Yang, Ling and Zheng, Jiayi and Wang, Heyuan and Liu, Zhongyi and Huang, Zhilin and Hong, Shenda and Zhang, Wentao and Cui, Bin},
  journal={IEEE Transactions on Knowledge and Data Engineering}, 
  title="{Individual and Structural Graph Information Bottlenecks for Out-of-Distribution Generalization}", 
  year={2024},
  volume={36},
  number={2},
  pages={682-693}
  }

@article{chromophore,
  title="{Experimental Database of Optical Properties of Organic Compounds}",
  author={Joung, Joonyoung F and Han, Minhi and Jeong, Minseok and Park, Sungnam},
  journal={Scientific Data},
  volume={7},
  number={1},
  pages={295},
  year={2020}
}

@article{mnsol,
  title="{Minnesota Solvation Database (MNSOL) Version 2012}",
  author={Marenich, Aleksandr V and Kelly, Casey P and Thompson, Jason D and Hawkins, Gregory D and Chambers, Candee C and Giesen, David J and Winget, Paul and Cramer, Christopher J and Truhlar, Donald G},
  year={2020}
}

@article{freesolv,
  title="{FreeSolv: A Database of Experimental and Calculated Hydration Free Energies, with Input Files}",
  author={Mobley, David L and Guthrie, J Peter},
  journal={Journal of Computer-Aided Molecular Design},
  volume={28},
  pages={711--720},
  year={2014}
}

@article{comsol,
  title="{Estimation of Solvation Quantities from Experimental Thermodynamic Data: Development of the Comprehensive CompSol Databank for Pure and Mixed Solutes}",
  author={Moine, Edouard and Privat, Romain and Sirjean, Baptiste and Jaubert, Jean-No{\"e}l},
  journal={Journal of Physical and Chemical Reference Data},
  volume={46},
  number={3},
  year={2017}
}

@article{abraham,
  title="{Mathematical Correlations for Describing Solute Transfer into Functionalized Alkane Solvents Containing Hydroxyl, Ether, Ester or Ketone Solvents}",
  author={Grubbs, Laura M and Saifullah, Mariam and Nohelli, E and Ye, Shulin and Achi, Sai S and Acree Jr, William E and Abraham, Michael H},
  journal={Fluid Phase Equilibria},
  volume={298},
  number={1},
  pages={48--53},
  year={2010}
}

@article{combisolv,
  title="{Transfer Learning for Solvation Free Energies: From Quantum Chemistry to Experiments}",
  author={Vermeire, Florence H and Green, William H},
  journal={Chemical Engineering Journal},
  volume={418},
  pages={129307},
  year={2021}
}

@article{zhangddi,
  title="{Predicting Potential Drug-Drug Interactions by Integrating Chemical, Biological, Phenotypic and Network Data}",
  author={Zhang, Wen and Chen, Yanlin and Liu, Feng and Luo, Fei and Tian, Gang and Li, Xiaohong},
  journal={BMC Bioinformatics},
  volume={18},
  pages={1--12},
  year={2017}
}

@article{hddi,
  title="{Heterogeneous Network Edge Prediction: A Data Integration Approach to Prioritize Disease-Associated Genes}",
  author={Himmelstein, Daniel S and Baranzini, Sergio E},
  journal={PLoS Computational Biology},
  volume={11},
  number={7},
  pages={e1004259},
  year={2015}
}

@article{drugbankddi,
  title="{DrugBank 5.0: A Major Update to the DrugBank Database for 2018}",
  author={Wishart, David S and Feunang, Yannick D and Guo, An C and Lo, Elvis J and Marcu, Ana and Grant, Jason R and Sajed, Tanvir and Johnson, Daniel and Li, Carin and Sayeeda, Zinat and others},
  journal={Nucleic Acids Research},
  volume={46},
  number={D1},
  pages={D1074--D1082},
  year={2018}
}

@article{manteltest,
  title="{Mantel Test in Population Genetics}",
  author={Diniz-Filho, Jos{\'e} Alexandre F and Soares, Thannya N and Lima, Jacqueline S and Dobrovolski, Ricardo and Landeiro, Victor Lemes and Telles, Mariana Pires de Campos and Rangel, Thiago F and Bini, Luis Mauricio},
  journal={Genetics and Molecular Biology},
  volume={36},
  pages={475--485},
  year={2013},
  publisher={SciELO Brasil}
}

@inproceedings{du2024mmgnnIJCAI,
  title="{MMGNN: A Molecular Merged Graph Neural Network for Explainable Solvation Free Energy Prediction}",
  author={Du, Wenjie and Zhang, Shuai and Di Wu, Jun Xia and Zhao, Ziyuan and Fang, Junfeng and Wang, Yang},
  booktitle={Proceedings of the Thirty-Third International Joint Conference on Artificial Intelligence},
  pages={5808--5816},
  year={2024}
}

@article{maboudi2024retuning,
  title="{Retuning of Hippocampal Representations During Sleep}",
  author={Maboudi, Kourosh and Giri, Bapun and Miyawaki, Hiroyuki and Kemere, Caleb and Diba, Kamran},
  journal={Nature},
  pages={1--9},
  year={2024},
  publisher={Nature Publishing Group UK London}
}

@article{xia2023understanding,
  title="{Understanding the Limitations of Deep Models for Molecular Property Prediction: Insights and Solutions}",
  author={Xia, Jun and Zhang, Lecheng and Zhu, Xiao and Liu, Yue and Gao, Zhangyang and Hu, Bozhen and Tan, Cheng and Zheng, Jiangbin and Li, Siyuan and Li, Stan Z},
  journal={Advances in Neural Information Processing Systems},
  volume={36},
  pages={64774--64792},
  year={2023}
}

@article{mcgibbon2024intuition,
  title="{From Intuition to AI: Evolution of Small Molecule Representations in Drug Discovery}",
  author={McGibbon, Miles and Shave, Steven and Dong, Jie and Gao, Yumiao and Houston, Douglas R and Xie, Jiancong and Yang, Yuedong and Schwaller, Philippe and Blay, Vincent},
  journal={Briefings in Bioinformatics},
  volume={25},
  number={1},
  pages={bbad422},
  year={2024},
  publisher={Oxford University Press}
}

@article{li2023reaction,
  title="{Reaction Performance Prediction with an Extrapolative and Interpretable Graph Model Based on Chemical Knowledge}",
  author={Li, Shu-Wen and Xu, Li-Cheng and Zhang, Cheng and Zhang, Shuo-Qing and Hong, Xin},
  journal={Nature Communications},
  volume={14},
  number={1},
  pages={3569},
  year={2023},
  publisher={Nature Publishing Group UK London}
}

@article{mak2024artificial,
  title="{Artificial Intelligence in Drug Discovery and Development}",
  author={Mak, Kit-Kay and Wong, Yi-Hang and Pichika, Mallikarjuna Rao},
  journal={Drug Discovery and Evaluation: Safety and Pharmacokinetic Assays},
  pages={1461--1498},
  year={2024},
  publisher={Springer}
}

@article{yang2024interaction,
  title="{Interaction-Based Inductive Bias in Graph Neural Networks: Enhancing Protein-Ligand Binding Affinity Predictions From 3D Structures}",
  author={Yang, Ziduo and Zhong, Weihe and Lv, Qiujie and Dong, Tiejun and Chen, Guanxing and Chen, Calvin Yu-Chian},
  journal={IEEE Transactions on Pattern Analysis and Machine Intelligence},
  year={2024},
  publisher={IEEE}
}

@inproceedings{TGNN,
author = {Seo, Sangwoo and Kim, Sungwon and Jung, Jihyeong and Lee, Yoonho and Park, Chanyoung},
title = "{Self-Explainable Temporal Graph Networks based on Graph Information Bottleneck}",
year = {2024},
isbn = {9798400704901},
booktitle = {Proceedings of the 30th ACM SIGKDD Conference on Knowledge Discovery and Data Mining},
pages = {2572–2583},
numpages = {12},
series = {KDD '24}
}

@article{koshland1995key,
  title="{The Key--Lock Theory and the Induced Fit Theory}",
  author={Koshland Jr, Daniel E},
  journal={Angewandte Chemie International Edition in English},
  volume={33},
  number={23-24},
  pages={2375--2378},
  year={1995},
  publisher={Wiley Online Library}
}

@article{stiller2022structure,
  title="{Structure Determination of High-Energy States in a Dynamic Protein Ensemble}",
  author={Stiller, John B and Otten, Renee and H{\"a}ussinger, Daniel and Rieder, Pascal S and Theobald, Douglas L and Kern, Dorothee},
  journal={Nature},
  volume={603},
  number={7901},
  pages={528--535},
  year={2022},
  publisher={Nature Publishing Group UK London}
}

@article{cao2024towards,
  title="{Towards Scalable Automated Alignment of LLMs: A Survey}",
  author={Cao, Boxi and Lu, Keming and Lu, Xinyu and Chen, Jiawei and Ren, Mengjie and Xiang, Hao and Liu, Peilin and Lu, Yaojie and He, Ben and Han, Xianpei and others},
  journal={arXiv preprint arXiv:2406.01252},
  year={2024}
}

@book{brown2015silico,
  title="{In Silico Medicinal Chemistry: Computational Methods to Support Drug Design}",
  author={Brown, Nathan},
  year={2015},
  publisher={Royal Society of Chemistry}
}

@article{smiles,
author = {Weininger, David},
title = "{SMILES, a Chemical Language and Information System. 1. Introduction to Methodology and Encoding Rules}",
year = {1988},
volume = {28},
number = {1},
issn = {0095-2338},
journal = {Journal of Chemical Information and Computer Sciences},
pages = {31–36},
numpages = {6}
}

@ARTICLE{TNNLS1,
  author={Huang, Han and Sun, Leilei and Du, Bowen and Lv, Weifeng},
  journal={IEEE Transactions on Neural Networks and Learning Systems}, 
  title={Learning Joint 2-D and 3-D Graph Diffusion Models for Complete Molecule Generation}, 
  year={2024},
  volume={35},
  number={9},
  pages={11857-11871},
  doi={10.1109/TNNLS.2024.3416328}}

@ARTICLE{TNNLS2,
  author={Chen, Mengjie and Zhang, Ming and Yan, Guiying and Wang, Guanghui and Qu, Cunquan},
  journal={IEEE Transactions on Neural Networks and Learning Systems}, 
  title={MRHGNN: Enhanced Multimodal Relational Hypergraph Neural Network for Synergistic Drug Combination Forecasting}, 
  year={2025},
  volume={36},
  number={9},
  pages={17086-17098},
  doi={10.1109/TNNLS.2025.3553385}}

@ARTICLE{TNNLS3,
  author={Lv, Qiujie and Chen, Guanxing and Yang, Ziduo and Zhong, Weihe and Chen, Calvin Yu-Chian},
  journal={IEEE Transactions on Neural Networks and Learning Systems}, 
  title={Meta-MolNet: A Cross-Domain Benchmark for Few Examples Drug Discovery}, 
  year={2025},
  volume={36},
  number={3},
  pages={4849-4863},
  doi={10.1109/TNNLS.2024.3359657}}

@inproceedings{zhang2025iterative,
  title={Iterative Substructure Extraction for Molecular Relational Learning with Interactive Graph Information Bottleneck},
  author={Zhang, Shuai and Fang, Junfeng and Li, Xuqiang and XIA, ALAN and Wei, Ye and Du, Wenjie and Wang, Yang and others},
  booktitle={The Thirteenth International Conference on Learning Representations},
  year={2025}
}

@ARTICLE{hu2024GIBsurvey,
  author={Hu, Shizhe and Lou, Zhengzheng and Yan, Xiaoqiang and Ye, Yangdong},
  journal={IEEE Transactions on Pattern Analysis and Machine Intelligence}, 
  title={A Survey on Information Bottleneck}, 
  year={2024},
  volume={46},
  number={8},
  pages={5325-5344}
  }

@inproceedings{zhang2026kdd,
  title="{Cross-Domain Molecular Relational Learning: Leveraging Chemical Structure-Activity Analysis}",
  author={Zhang, Peiliang and Yuan, Jingling and Wu, Shiqing and Hu Mengqing and Che, Chao and Zhu, Yongjun and Li, Lin},
  booktitle={Proceedings of the 32nd SIGKDD Conference on Knowledge Discovery and Data Mining},
  year={2026}
}

@article{zhang2026prototype,
  title={Prototype Learning with Structural-Semantic Alignment for Interpretable Molecular Relational Learning},
  author={Zhang, Peiliang and Yuan, Jingling and Wang, Jianmin and Zhu, Yongjun and Li, Lin},
  journal={Knowledge-Based Systems},
  pages={115460},
  year={2026},
  publisher={Elsevier}
}

@inproceedings{zhang2024key,
  title={Key substructure learning with chemical intuition for material property prediction},
  author={Zhang, Peiliang and Yuan, Jingling and Li, Lin and Luo, Wen and Hu, Jiwei and Li, Xin},
  booktitle={International Conference on Database Systems for Advanced Applications},
  pages={87--103},
  year={2024},
  organization={Springer}
}

@inproceedings{chi,
author = {Liu, Xuan and Shang, HaoYang and Jin, Haojian},
title = {CoBRA: Programming Cognitive Bias in Social Agents Using Classic Social Science Experiments},
year = {2026},
isbn = {9798400722783},
publisher = {Association for Computing Machinery},
address = {New York, NY, USA},
url = {https://doi.org/10.1145/3772318.3790804},
doi = {10.1145/3772318.3790804},
booktitle = {Proceedings of the 2026 CHI Conference on Human Factors in Computing Systems},
articleno = {64},
numpages = {30},
}

\appendix

\setcounter{table}{0}
\renewcommand{\thetable}{S\arabic{table}}
\setcounter{figure}{0}
\renewcommand{\thefigure}{S\arabic{figure}}

\section{Proof}
\subsection{The Proof of Theorem \ref{theorem 1}}
\label{The Proof of Theorem 1}
Since $\mathcal{P}(\mathcal{G}^n_x; \mathcal{Y}|\mathcal{G}^c_x)$ is non-negative and $\mathcal{G}_x^n$, $\mathcal{G}_y^n$ have little effect on $\mathcal{Y}$, according to the law of conditional probability, $\mathcal{P}({\mathcal{G}_{x}},{\mathcal{G}_{y}};\mathcal{Y})$ is expressed as:
\begin{equation}
\begin{aligned}
    \mathcal{P}({\mathcal{G}_{x}},{\mathcal{G}_{y}};\mathcal{Y}) &= \mathcal{P}(\mathcal{G}_x^c;\mathcal{Y})+\mathcal{P}(\mathcal{G}_y^n;\mathcal{Y}|\mathcal{G}_x^c,\mathcal{G}_x^n,\mathcal{G}_y^c)\\
    &+\mathcal{P}(\mathcal{G}^n_x; \mathcal{Y}|\mathcal{G}^c_x)+\mathcal{P}(\mathcal{G}_y^c;\mathcal{Y}|\mathcal{G}_x^c,\mathcal{G}_x^n,)\\
    &\geq \mathcal{P}(\mathcal{G}_x^c;\mathcal{Y})+\mathcal{P}(\mathcal{G}^c_y; \mathcal{Y}|\mathcal{G}^c_x)
\end{aligned}   
\end{equation}
$\mathcal{P}(\mathcal{G}^c_y;\mathcal{Y})$ represents the probability between $\mathcal{G}^c_y$ and $\mathcal{Y}$, which contains at least the conditional probability $\mathcal{P}({\mathcal{G}^{c}_y};\mathcal{Y}|{\mathcal{G}^{c}_c})$. We can obtain $\mathcal{P}({\mathcal{G}_{x}},{\mathcal{G}_{y}};\mathcal{Y}) \geq \mathcal{P}({\mathcal{G}^{c}};\mathcal{Y}|{\mathcal{G}^{n}})$. According to the properties of conditional probability:
\begin{equation}
    \begin{aligned}
        \mathcal{P}({\mathcal{G}^{c}};\mathcal{Y}|{\mathcal{G}^{n}}) &:= \mathcal{P}(\mathcal{G}_x^c;\mathcal{Y})+\mathcal{P}(\mathcal{G}^c_y; \mathcal{Y}|\mathcal{G}^c_x)\\
        &= \mathcal{P}(\mathcal{G}_x^c;\mathcal{Y})+\mathcal{P}(\mathcal{G}^c_y; \mathcal{Y})=\mathcal{P}(\mathcal{G}_x^c,\mathcal{G}^c_y;\mathcal{Y})
    \end{aligned}
    \label{eq2}
\end{equation}

Based on Eq (\ref{eq2}), and considering that $\mathcal{P}(\mathcal{G}_x^c;\mathcal{G}_x^n)$ and $\mathcal{P}(\mathcal{G}_y^c;\mathcal{G}_y^n)$ are non-negative, we have the equation as:
\begin{equation}
    \mathcal{P}({\mathcal{G}_{x}},\!{\mathcal{G}_{y}};\mathcal{Y})\!-\!\mathcal{P}(\mathcal{G}_x^c,\!\mathcal{G}^c_y;\!\mathcal{Y})\!+\!\mathcal{P}(\mathcal{G}_x^c;\!\mathcal{G}_x^n)\!+\!\mathcal{P}(\mathcal{G}_y^c;\!\mathcal{G}_y^n)\!\!\geq\!\! 0
    \label{eq3}
\end{equation}

Therefore, by increasing $\mathcal{P}(\mathcal{G}_x^c,\mathcal{G}^c_y;\mathcal{Y})$ and decreasing $\mathcal{P}(\mathcal{G}_x^c;\mathcal{G}_x^n)+\mathcal{P}(\mathcal{G}_y^c;\mathcal{G}_y^n)$ (\textit{i.e.}, increasing the correlation between the core substructures captured by the model and prediction targets, and decreasing the impact of the confounding substructures on the core substructure representation), we can ensure that there exists a minimal positive number $\varepsilon $ that satisfies the following relationship:
\begin{equation}
\left|\! \mathcal{P}({\mathcal{G}_{x}},{\mathcal{G}_{y}};\mathcal{Y})\!\!-\!\!\mathcal{P}(\mathcal{G}_x^c,\mathcal{G}^c_y;\mathcal{Y})\!\!+\!\!\mathcal{P}(\mathcal{G}_x^c;\mathcal{G}_x^n)\!\!+\!\!\mathcal{P}(\mathcal{G}_y^c;\mathcal{G}_y^n) \!\right|\! \le \! \varepsilon 
\end{equation}

\subsection{The proof of Eq (\ref{eq:op1})}
\label{The proof of Eq op1}
The $\mathcal{I}(\mathcal{G}^c_x,\mathcal{G}_x)$ is expressed in integral form as:
\begin{equation}
\label{eq:op1-1}
\begin{aligned}
\mathcal{I}(\mathcal{G}^c_x,\mathcal{G}_x)&=\int \!\!\!\! \int p(\mathcal{G}^c_x,\mathcal{G}_x)\log(\frac{p(\mathcal{G}^c_x,\mathcal{G}_x)}{p(\mathcal{G}^c_x)p(\mathcal{G}_x)})d\mathcal{G}^c_xd\mathcal{G}_x
\end{aligned}
\end{equation}

The probability function $\gamma$ in Eq (\ref{align}) is utilized to adjust the conditional probability distribution $p(\mathcal{G}^c_x,\mathcal{G}_x)$, and $\mathcal{G}^c_x$, $\mathcal{G}^n_x$ are conditionally independent of $\mathcal{G}_x$. We can obtain $p(\mathcal{G}^c_x|\mathcal{G}_x)=p(\mathcal{G}^c_x|\mathcal{G}_x,\gamma)$.
Further, $\mathcal{I}(\mathcal{G}^c_x,\mathcal{G}_x)$ is expressed as:
\begin{equation}
\begin{aligned}
\mathcal{I}(\mathcal{G}^c_x,\mathcal{G}_x)&\!\!=\!\!\int \!\!\!\! \int \!\! p(\mathcal{G}^c_x|\mathcal{G}_x, \! \gamma) \log(\frac{p(\mathcal{G}^c_x|\mathcal{G}_x, \! \gamma)}{p(\mathcal{G}^c_x)}) d\mathcal{G}_x^cd\mathcal{G}_x\\
&:= {KL}(p(\mathcal{G}^c_x|\mathcal{G}_x, \! \gamma)||p(\mathcal{G}^c_x))\\
& = \mathcal{L}_{KL_x}^c
\end{aligned}
\label{eq:op1-2}
\end{equation}

The result of Eq (\ref{eq:op1-2}) indicates that $\mathcal{I}(\mathcal{G}^c_x,\mathcal{G}_x)$ can be computed by the KL divergence between $\mathcal{G}^c_x$ and $\mathcal{G}_x$. Similarly, $\mathcal{I}(\mathcal{G}^n_x,\mathcal{G}_x)={KL}(p(\mathcal{G}^n_x|\mathcal{G}_x,\gamma)||p(\mathcal{G}^n_x))=\mathcal{L}_{KL_x}^n$.

Finally, we have the following equation as:
\begin{equation}
    \mathcal{I}(\mathcal{G}^c_x,\mathcal{G}^n_x)\le \min (\mathcal{L}_{KL_x}^c,\mathcal{L}_{KL_x}^n)
\end{equation}

\subsection{The proof of Eq (\ref{eq:op2})}
\label{The proof of Eq op2}
For the term $\mathcal{I}_{ca}=\mathcal{I}(\mathcal{Y};\mathcal{G}_{x}^{c},\mathcal{G}_{y}^{c})$, by definition:
\begin{equation}
    \mathcal{I}_{ca} \!\!=\!\!\!\int\!\!\!\!\!\int\!\!\!\!\!\int\!\!\! p(\mathcal{Y},\mathcal{G}_{x}^{c},\mathcal{G}_{y}^{c})\! \log (\frac{p(\mathcal{Y},\mathcal{G}_{x}^{c},\mathcal{G}_{y}^{c})}{p(\mathcal{Y})})d\mathcal{Y}d\mathcal{G}_{x}^{c}d\mathcal{G}_{y}^{c}
\end{equation}

We introduce the variational approximation distribution $q(\mathcal{G}_{x}^{c},\mathcal{G}_{y}^{c}|\mathcal{Y})$ to approximate the conditional probability distribution $p(\mathcal{G}_{x}^{c},\mathcal{G}_{y}^{c}|\mathcal{Y})$. Then, the $\mathcal{I}_{ca}$ is expressed as:
\begin{equation}
\begin{aligned}
    \mathcal{I}_{ca}\!\!&
    = \!\!\!\!\int\!\!\!p(\mathcal{Y})\!\!\!\int\!\!\!\!\! \int\!\!\! q(\mathcal{G}_{x}^{c},\mathcal{G}_{y}^{c}|\mathcal{Y})\log (\frac{p(\mathcal{Y},\mathcal{G}_{x}^{c},\mathcal{G}_{y}^{c})}{q(\mathcal{G}_{x}^{c},\mathcal{G}_{y}^{c})})d\mathcal{Y}d\mathcal{G}_{x}^{c}d\mathcal{G}_{y}^{c}\\
    &=\!\!\!\!\int\!\!\!p(\!\mathcal{Y}\!)\!\!\!\int\!\!\!\!\! \int\!\!\! q(\mathcal{G}_{x}^{c},\!\mathcal{G}_{y}^{c}|\mathcal{Y}\!)\!\log (\!\frac{p(\mathcal{Y})p(\mathcal{G}_{x}^{c}\!,\!\mathcal{G}_{y}^{c}|\mathcal{Y})}{q(\mathcal{G}_{x}^{c},\!\mathcal{G}_{y}^{c})}\!)d\mathcal{Y}d\mathcal{G}_{x}^{c}d\mathcal{G}_{y}^{c}\\
    &=\!\!\!\!\int\!\!p(\mathcal{Y})(\log p(\mathcal{Y})+\!\!\!\int\!\!\!\!\! \int\!\!{\mathcal{F}_{ca}(\mathcal{G}_{x}^{c},\mathcal{G}_{y}^{c}|\mathcal{Y})}d\mathcal{G}_{x}^{c}d\mathcal{G}_{y}^{c})d\mathcal{Y}\\
    &=\!\!\!\!\int\!\!p(\mathcal{Y})\log p(\mathcal{Y})d\mathcal{Y}\\
    &+\!\!\!\!\int\!\!p(\mathcal{Y})\!\!\!\int\!\!\!\!\! \int\!\!{\mathcal{F}_{ca}(\mathcal{G}_{x}^{c},\mathcal{G}_{y}^{c}|\mathcal{Y})}d\mathcal{G}_{x}^{c}d\mathcal{G}_{y}^{c})d\mathcal{Y}\\
    &=\!H(\mathcal{Y})\!\!+\!\!\!\int\!\!{p(\mathcal{Y})\!\!\iint\!\!{{\mathcal{F}_{ca}(\mathcal{G}_{x}^{c},\mathcal{G}_{y}^{c}|\mathcal{Y})}}d\mathcal{G}_{x}^{c}d\mathcal{G}_{y}^{c}d\mathcal{Y}}
\end{aligned}
\end{equation}

$H(\mathcal{Y})$ is the entropy of $\mathcal{Y}$. Finally, we have the following equation as:
\begin{equation}
\mathcal{I}(\mathcal{Y};\!\mathcal{G}_{x}^{c},\!\mathcal{G}_{y}^{c})\!\ge\!\! \int\!\!{p(\mathcal{Y})\!\!\int\!\!\!\!\!\int\!\!{{\mathcal{F}_{ca}(\mathcal{G}_{x}^{c},\mathcal{G}_{y}^{c}|\mathcal{Y})}}d\mathcal{G}_{x}^{c}d\mathcal{G}_{y}^{c}d\mathcal{Y}}
\end{equation}

\section{Results Analysis}

\subsection{Ablation Experiment}
\begin{figure}[htpb]
\centering
\includegraphics[width=\columnwidth]{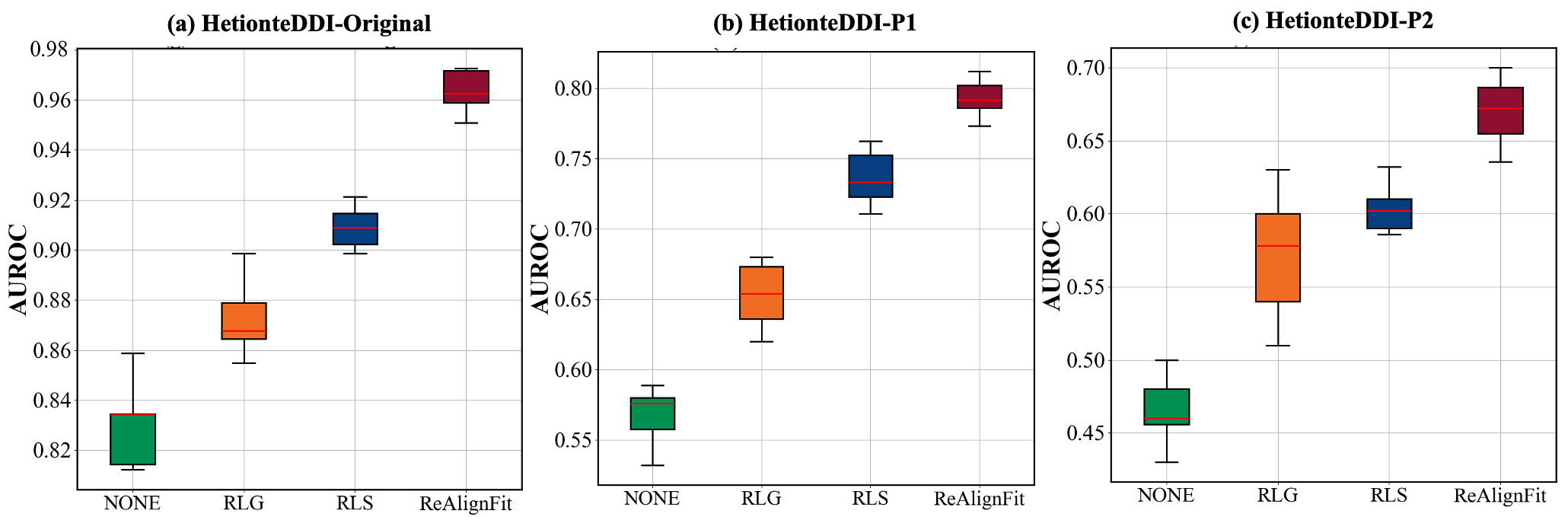} 
\caption{The experimental results of ablation experiment.}
\label{aba}
\end{figure}

\begin{figure}[htpb]
\centering
\includegraphics[width=\columnwidth]{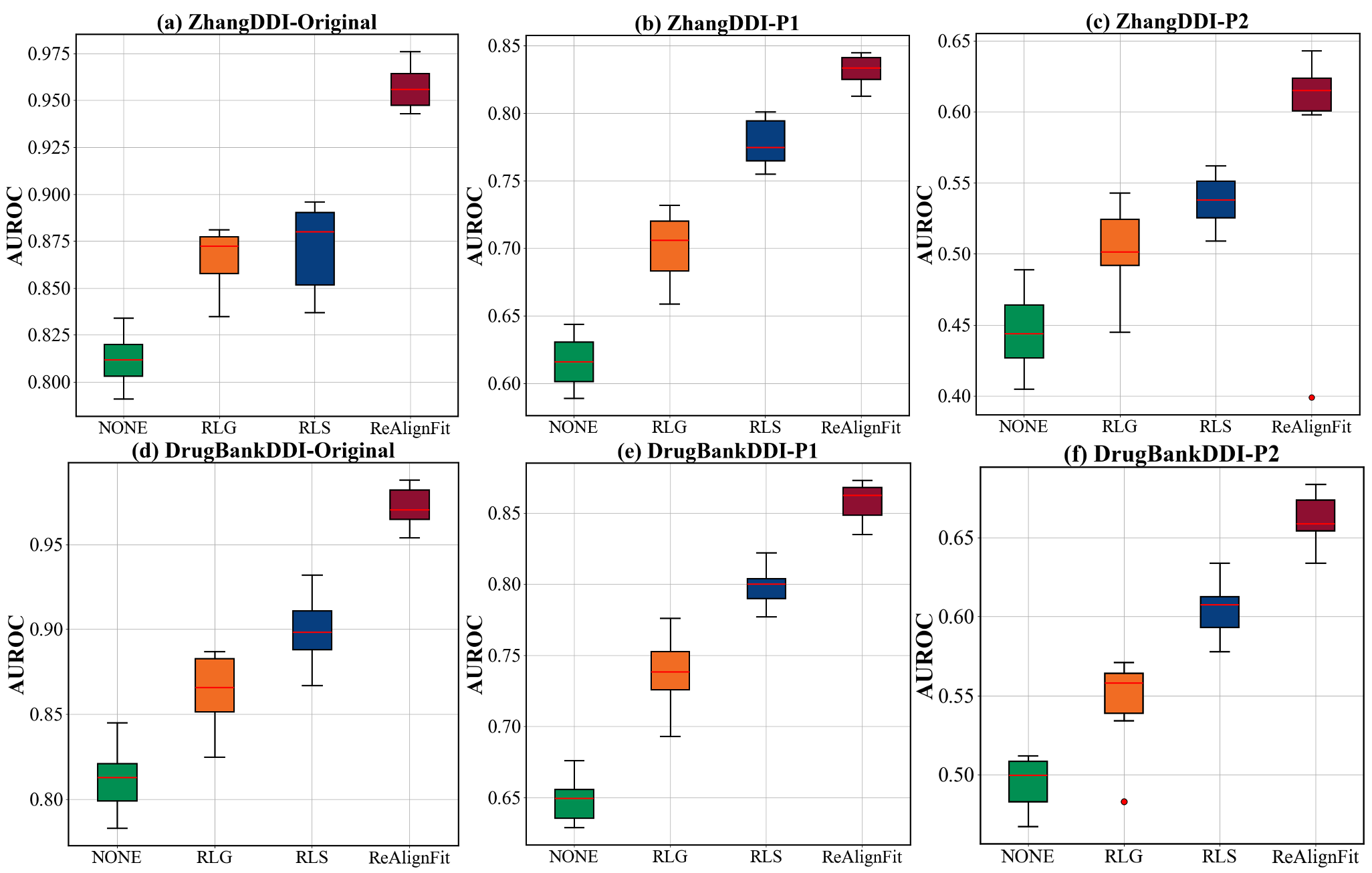} 
\caption{The experimental results of ablation experiment.}
\label{aba-supp}
\end{figure}

To analyze the impact of representational alignment and S-GIB on the overall performance of the model, we compared the following model variants: NONE (no representational alignment and optimization), RLG (optimized substructure representation based on S-GIB), and RLS (optimized substructure representation based on representational alignment). The results are shown in Figs. \ref{aba} and \ref{aba-supp}.

\begin{itemize}
    \item \textbf{NONE}: No representational alignment is applied. Relationship prediction relies solely on substructures extracted by SRIN as molecular feature representations.
    \item \textbf{RLG}: S-GIB is used for optimizing substructure representations, focusing on molecular-level alignment and assessing S-GIB's impact on MRL stability without representational alignment.
    \item \textbf{RLS}: Representational alignment is performed at the substructure level, modeling dynamic interactions between substructures for MRL, excluding S-GIB's optimization. 
\end{itemize}

The experimental results in Figs. \ref{aba} and \ref{aba-supp} demonstrate that incorporating S-GIB optimization and substructure representational alignment improves AUROC by 4\%, 8\%, and 10\%, and by 8\%, 16\%, and 13\%, respectively, for the three data distributions in the HetionetDDI dataset. Similar trends are observed in the ZhangDDI and DrugBankDDI datasets. Notably, the performance gains in AUROC achieved by substructure representational alignment are higher than those from S-GIB optimization. These experimental results suggest that representational alignment and S-GIB optimization are critical critical to enhancing ReAlignFit's predictive performance.

Comparing the distribution ranges of AUROC of the five ablation experiments in Figs. \ref{aba} and \ref{aba-supp}, we observe that RLS variant exhibits tighter AUROC ranges on distribution-shifted data. This result indicates that substructure representational alignment effectively enhances model stability. In contrast, the contribution of S-GIB optimization to stability is negligible and even reduces the model stability in individual scenarios. These results suggest that selecting substructure pairs with high functional compatibility through representational alignment is the key to stability improvement.

Overall, representational alignment improves MRL stability by identifying highly compatible binding sites. S-GIB enhances predictive performance by mitigating the influence of confounding substructures on highly compatible substructures. The synergy between representational alignment and S-GIB achieves a win-win situation regarding model predictive performance and stability.

\subsection{ Confusion Analysis on $\alpha$ and $\beta$}
\begin{figure}[t]
    \centering
    \includegraphics[width=\columnwidth]{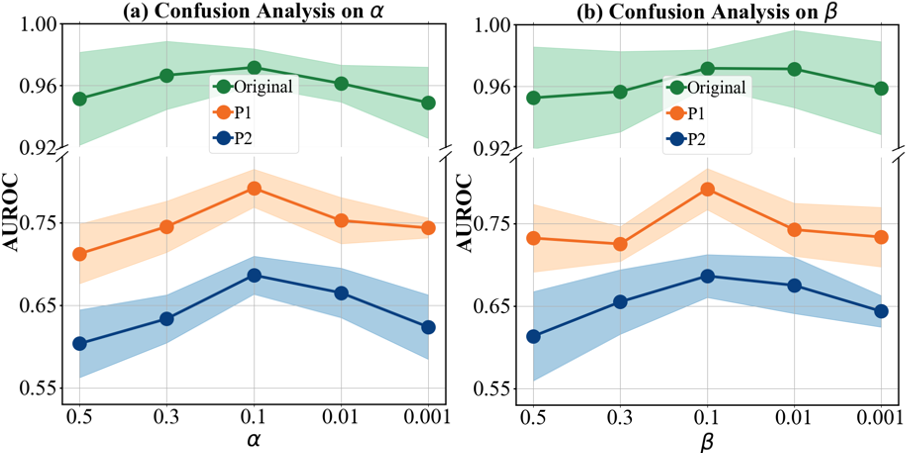}
    \caption{The experimental results of confusion analysis in HetionteDDI dataset.}
    \label{fig:confusion-result}
\end{figure}

To verify the impact of confounding information $\alpha I(G_{x}^{c},G_{x}^{n})+ \beta I(G_{y}^{c},G_{y}^{n})$ in Eq (\ref{eq:CSGIBnew}) on the stability of ReAlignFit's predictive performance, we set $\alpha$ and $\beta$ to $\{0.001,0.01,0.1,0.3,0.5\}$ and conduct relevant experiments in three different data distributions. The experimental results are shown in Fig. \ref{fig:confusion-result}.

In Fig. \ref{fig:confusion-result}, there exists an optimal value of $\alpha$ and $\beta$ (i.e., $\alpha \!\!=\!\! \beta \!\!= \!\!0.1$) that balances the model's predictive performance and the impact of confounding information on molecular representation. When $\alpha$ and $\beta$ are excessively large ($\alpha \!\!= \!\!\beta \!\!\ge \!\!0.3$), the model's predictive performance fluctuates significantly in different data distributions, and its stability decreases markedly. This occurs because high values of $\alpha$ and $\beta$  cause the model to learn excessive confounding information, making information difficult to use in MRL effectively. Conversely, when $\alpha$ and $\beta$ are smaller ($\alpha \!\!=\!\! \beta \!\!< \!\!0.1$), while the model's performance fluctuations are reduced, its predictive performance remains below optimal levels. In this case, the model struggles to adequately distinguish between confounding and core substructures, consequently affecting its performance and generalization. Therefore, appropriately adjusting $\alpha$ and $\beta$ enables the model to more effectively balance the impact of confounding and core information on molecular representation to improve the stability of the model's prediction performance.

\end{document}